\pdfoutput=1

\documentclass[11pt]{article}

\usepackage{EMNLP2023}

\usepackage{times}
\usepackage{latexsym}

\usepackage{amsmath}
\usepackage{amssymb}
\usepackage{mathtools}
\usepackage{amsthm}
\usepackage{amsfonts}

\usepackage{graphicx}
\usepackage{multirow}
\usepackage{makecell}
\usepackage{booktabs} 

\usepackage[T1]{fontenc}

\usepackage[utf8]{inputenc}

\usepackage{microtype}

\usepackage{inconsolata}

\title{LagKV: Lag-Relative Information of the KV Cache Tells Which Tokens Are Important}

\author{Manlai Liang \and Jiaming Zhang \and Xiong Li \and Jinlong Li\footnote{Corresponding Author} \\
 AI Lab, China Merchants Bank, China \\
  \texttt{\{liangml,zhangjm,lixiong,lucida\}@cmbchina.com} \\}

\begin{document}
\maketitle

\begin{abstract}
	The increasing size of the Key-Value (KV) cache during the Large Language Models long-context inference is the main obstacle for its balance between the deployment cost and task accuracy.
	To reduce the KV cache size in such scenarios, most previous efforts leveraged on the attention weight to evict non-critical cache tokens. 
	But there is a trade-off in those methods, they usually require major modification of the inference infrastructure and significant computation overhead.
	Based on the fact that the Large Language models are autoregressive models, we propose LagKV, a KV compression strategy only relying on straight forward comparison among KV themselves. It is a totally attention free method which offers easy integration to the main stream inference platform and comparable performance comparing to other complicated KV compression methods. Results on RULER benchmark show that, our approach outperforms SnapKV and StreamingLLM in different compression ratios. Especially in the 64-digit passkey retrieval task, our method outperforms the attention weight based method $H_2O$ over $50\%$ with same compression ratios.
Our code is available at \url{https://github.com/AI-Lab-China-Merchants-Bank/LagKV}.
\end{abstract} 
\section{Introduction}
Large Language Models (LLMs) have recently demonstrated remarkable success across diverse text processing tasks, including document retrieval~\cite{laban2023summedits}, code generation~\cite{gu2023llm}, and mathematical reasoning (like R1 model~\cite{deepseekai2025deepseekr1incentivizingreasoningcapability}). The Scaling law~\cite{kaplan2020scalinglawsneurallanguage} suggests that larger models generally achieve superior performance. The R1-like models further indicates that longer generation sequences with additional 'thinking tokens' can enhance reasoning capabilities. However, these improvements comes at a significant cost: the growing KV cache size poses a major challenge for efficient LLM inference. Many efforts try to mitigate this challenge.

Most of LLMs are totally relying on Self-Attention mechanism~\cite{vaswani2023attentionneed} to determine which historical tokens are important in the next token prediction. Therefore, many KV compression approaches are based on it to drop unimportant ones ~\citep{H2o,liu2024scissorhands,SnapKV, questpub, kvpress}. This kind of algorithms keeps a remarkable performance even when the compression ratio is high. However, most of these importance-based token-dropping approaches depend on the ending query question (Instruction Dependence) to achieve such a performance ~\cite{li2025scbench, AdaKV, razorattention}. 

Another prominent direction in KV cache optimization involves quantization techniques ~\citep{PyramidInfer,liu2024kivi}, which aim to compress the memory footprint of KV states by representing them with reduced precision. These methods achieve significant memory savings—often by 4× or more—while preserving model performance through careful error mitigation strategies. Beyond memory efficiency, quantization also reduces the bandwidth overhead of transferring KV cache across devices in distributed inference scenarios, accelerating multi-GPU or memory-bound workloads. However, a critical limitation of pure quantization approaches is that they retain all historical tokens, leaving the computational cost of attention unchanged. For long-context tasks, this means the quadratic complexity of attention persists despite the reduced memory usage. 

The simple but with limited performance methods are usually based on the sliding window tokens eviction. Sliding window-based eviction methods—such as those used in Infinite-LLM~\cite{han2024lminfinitezeroshotextremelength} and StreamingLLM~\cite{SLM}—retain only the initial cache tokens and those within a fixed sliding window, discarding the rest. However, this indiscriminate eviction strategy often leads to a notable degradation in generation quality.

Recent work by ~\citep{cachegen2024,liu2024kivi} addresses the statistical properties of KV states, revealing distinct distribution patterns for keys and values. Their findings suggest that per-channel quantization for keys (which exhibit consistent variance across feature dimensions) and per-token quantization for values (which vary more significantly across sequence positions) yield better fidelity. 
This observation motivates our key insight: token importance for eviction—traditionally derived from attention weights—can instead be inferred from token- and channel-wise distribution patterns in the KV space. By leveraging these structural properties, we can design a pruning criterion, LagKV, that is both hardware-friendly (compatible with Flash Attention~\cite{dao2023flashattention}) and instruction independent, enabling compute savings alongside memory reduction.

\begin{figure*}[t]
	\centering
	\includegraphics [width=1.0\textwidth]{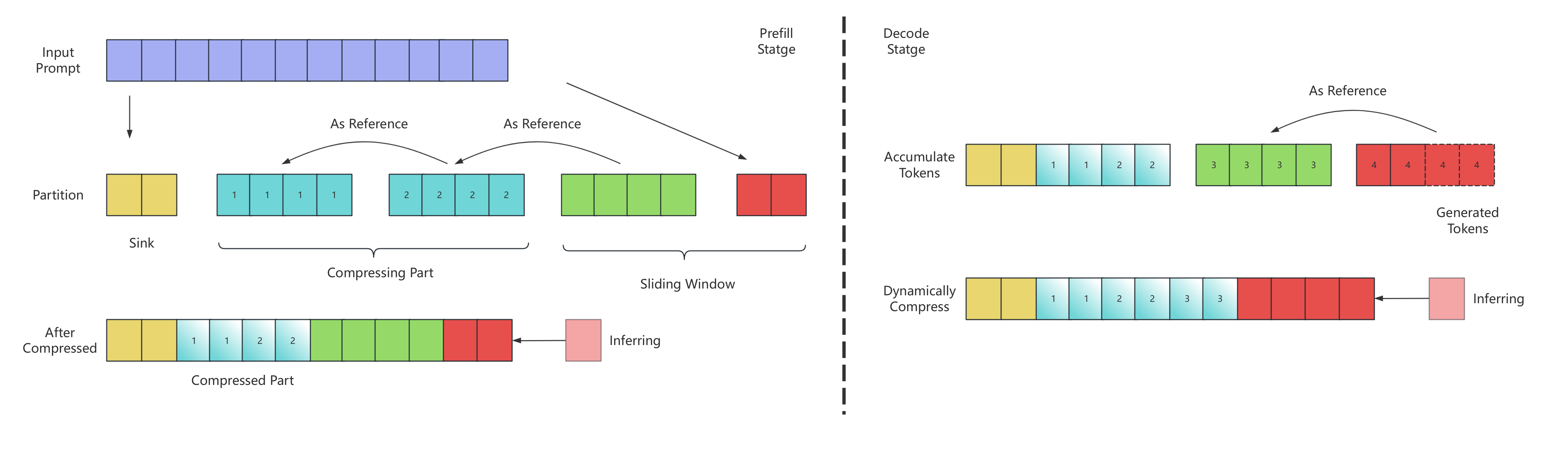}
		\vspace{-1cm}
	\caption{LagKV recursively compression process: partition the KV cache and use the next joint chunk as reference to compress the current one. Keep the rest of them as the sliding window.}		
	\label{fig:lag_kv_flow}
	\vspace{-2pt}
\end{figure*}

\section{Methodology}
In this section, we formally introduce our KV compression method, LagKV. We begin by looking at the autoregressive process of the LLMs. Inspired by this, we propose a simple yet effective strategy to use the subsequent tokens to compress the previous ones.

\subsection{Preliminaries}
LLMs' next token prediction relies on the previous tokens. First, in the prefill stage, the model uses its tokenizer to convert the words to $n$ indices of the embedding metrics $E\in \mathbb{R}^{V \times d}$ of the model and collects the representations to form a input matrix, $X\in \mathbb{R}^{n \times d}$. This matrix is the initial tokens of the first layer of LLM and then each layer will output a same shape matrix as next layer's input.
To depict the operations in each layer, we follow the notation system from ~\cite{liu2023dejavucontextualsparsity} with $h$ attention heads.
For each head $i \in [1, h]$ and head dimension $d_h$, we focus on the Query, Key, and Value states, which are converted from tokens by three linear transformation matrices $W_i^Q$, $W_i^K$, $W_i^V \in \mathbb{R}^{d \times d_h}$ separately:
\begin{equation}
	Q_i=XW_i^Q,K_i = XW_i^K,V_i=XW_i^V
\end{equation}
The output $Y \in \mathbb{R}^{n \times d}$ is computed using the attention weights $A_i \in \mathbb{R}^{n \times n}$ and  the final output matrix $W^O \, \in \mathbb{R}^{d \times d}$:

\begin{equation}
	Y = Concat_{i \in [1, h]}( A_i V_i )W^O
\end{equation}
where
\begin{equation}
 A_i =\text{softmax}(\frac{Q_{i}K_i^{T}}{\sqrt{d_h}} ).
\end{equation}
 When the new tokens are generated subsequently in the  autoregressive inference, which named as decode stage, the embedding of generated token $x$ is mapped to its respective  Query, Key, and Value states for each head, and the previous KV cache is updated accordingly:
\begin{equation}
	q_i = xW_i^Q,k_i = xW_i^K,v_i = xW_i^V
\end{equation}
\begin{equation}
	K_i = Cat[K_i:k_i],V_i=Cat[V_i:v_i]
\end{equation}
\begin{equation}
A_i =\text{softmax}(\frac{q_{i}K_i^{T}}{\sqrt{d_h}} )
\end{equation}
Since $q_i \in \mathbb{R}^{1 \times d_h}$, the computation will be much faster because of the KV cache.

\subsection{LagKV}
Since the intrinsic property of autoregressive model, the next token representation will not change abruptly from the previous one. As observed in \cite{cachegen2024}, the called token-wise locality will show that the tokens in closer proximity have more similar K/V tensor values compared to tokens that are further apart. 

And also, the StreamingLLM method \cite{SLM} has demonstrated that the head portion and sliding window of the KV cache are crucial. This suggests that cache compression should use subsequent tokens to assess whether prior tokens remain in the cache, rather than relying on the competition between them—as done in many attention-weight-based methods.

Inspired by above insights, we proposed our LagKV method as:
\begin{itemize}
\item After the prefill is done, start to apply the compression dynamically.

\item Always keep the attention sink with size $S$ and the already compressed part if had unchanged.

\item Skip the compression if the length of the rest KV after the static part is less than $2L$, where we denote the lag size as $L$. 

\item Partition the rest KV with $L$. If it's not divisible by $L$, the modulo of it will be added to the sliding window.

\item Recursively compute the KV cache score. Use the next partition as a reference, calculate token-wise max and min from the reference then use max-min to normalize the Key and Value states respectively. After the KV are normalized, calculate the channel-wise standard deviation then softmax. The equations are formally like:
\begin{equation}
min_i^{p,Z} = min_{seq}({Z_i^{p+1}})
\label{eq:min}
\end{equation}
\begin{equation}
max_i^{p,Z}=max_{seq}({Z_i^{p+1}})
\label{eq:max}
\end{equation}
\begin{equation}
\bar{Z_i^p}=\frac{{Z_i^p}-min_i^{p,Z}}{max_i^{p,Z}-min_i^{p,Z}}
\end{equation}
\begin{equation}
score(Z_i)=Softmax(Std.(\bar{Z_i}))
\end{equation}
where $Z$ is one of $\{K, V\}$, $p$ denotes the partition index, $i$ represents the head index and $seq$ for the sequence axis. Since the last partition has no reference can be used, our method will naturally have a sliding window with at least size $L$.

\item Sum the scores of Key and Value to get the final score of each token:
\begin{equation}
score_{i}=score(K_i)+score(V_i)
\label{eq:scoresum}
\end{equation}

\item Base on the $score_i$, use the top-K strategy to select tokens in each partition and each head and add them to the compressed part.
\end{itemize}

The max-min normalization is applied along the sequence dimension, meaning each channel is normalized using statistics from lag-$L$ tokens. Due to token-wise locality, the channel-specific norms of $K_i$ and $V_i$ are largely eliminated. The resulting normalized representations, $\bar{K_i}$ and $\bar{V_i}$, retain the original channel-wise variance, allowing the standard deviation to serve as a measure of token importance. The softmax operation then identifies and separates outliers, while the summed scores 
$score(K_i)$ and $score(V_i)$ determine their relative contributions.

As showed in Fig.~\ref{fig:lag_kv_flow}, our method is recursively compressing KV cache in both prefill and decode parts, which is essential for the token-wise locality as mentioned above. It requires relative short distance to keep the similarity among the KV states.
Subsequently, another benefit, it also avoids the bias from the long context with length much larger than $L$ and the case when the question is at the end of the prompt.

We do not compare the LagKV score to the attention weights here. The attention weights vary on different incoming queries. But our scoring method does not depend on the query states or the tokens after the next joint chunk. It mainly finds the tokens that are not coherent to the next chunk and keep them in the cache. As in KIVI~\cite{liu2024kivi} quantization method, we need a rightful mean to find the correct variance and then prune the small ones. However, we use this strategy to evict tokens instead of quantizing them.

To calculate the compression ratio, we set the token retention ratio as $r$ in each partition. In the partition chunk, only $rL$ tokens will be kept and others are evicted. Therefore, the compression ratio $C$ for the token sequence length $L_s \ge S+2L$ can be expressed as:
\begin{equation}
L_R=S+rL (\lfloor\frac{L_s-S}{L}\rfloor - 1) + L + Mod(L_s-S, L)
\end{equation}
\begin{equation}
\label{eq:compress_ratio}
C=1-\frac{L_R}{L_s}
\end{equation}
Where $L_R$ is the length of the KV cache after compression. For the case $L_s < S+2L$, the compression ratio is zero.

\section{Comparisons}
\subsection{Base Models}
\label{sec:settings}
We employ two open-source base models: Llama-3.1-8B-Instruct~\cite{llama3} and Qwen2.5-7B-Instruct~\cite{qwen2025qwen25technicalreport}. These models are main stream LLMs with moderate size and both leverage the GQA~\cite{GQA} technique to reduce the KV cache size.

\subsection{Results of RULER}

We use the RULER~\cite{hsieh2024ruler} benchmark to compare our approach to SnapKV~\cite{SnapKV} and StreamingLLM~\cite{SLM} in various compression ratios.
To fairly compare different methods, we integrate our approach into the framework KVPress~\cite{kvpress} and adapt their versions of other approaches. This framework applies compression without question to avoid the query-aware bias. In this task, we set the lag size to be $L=128$ for LagKV and the retention ratio of each recursive window will be adaptively changed by Eq. \ref{eq:compress_ratio} for different compression ratios.

The results are present in Table~\ref{ruler_llama} and ~\ref{ruler_qwen} with best scores of each compression ratio shown in bold. The average scores of RULER tasks show that LagKV outperforms SnapKV and StreamingLLM across all compression ratios.

\begin{table*}[ht!]
\centering
\caption{RULER-16K Results of Llama-3.1-8B-Instruct}
\label{ruler_llama}
\resizebox{\textwidth}{!}{
\begin{tabular}{c|c|ccccccccccccc|c}
\toprule
\textbf{Comp. Ratio}   & Method & SK1 & SK2 & SK3 & MK1 & MK2 & MK3 & MV & MQ & VT & CWE & FWE & QA1 & QA2 & AVERAGE \\
\midrule
\multirow{1}{*}{0.0}
& FullKV & 100.0 & 100.0 & 100.0 & 97.4 & 100.0 & 100.0 & 100.0 & 98.2 & 100.0 & 90.2 & 87.5 & 75.7 & 54.7 & 92.6\\
\midrule
\multirow{3}{*}{0.25}
& SnapKV & \textbf{100.0} & \textbf{100.0} & 33.3 & \textbf{98.7} & 83.3 & 63.9 & 97.9 & 98.2 & 94.8 & 85.3 & \textbf{90.2} & 64.9 & 46.9 & 81.3\\
& StreamingLLM & 72.5 & 74.7 & 72.5 & 79.2 & 86.7 & \textbf{66.7} & 72.7 & 75.0 & 90.5 & 0.1 & 87.1 & \textbf{75.7} & 43.8 & 69.0\\
& LagKV & \textbf{100.0} & \textbf{100.0} & \textbf{95.7} & 97.4 & \textbf{96.7} & 56.9 & \textbf{99.4} & \textbf{98.5} & \textbf{100.0} & \textbf{88.4} & 89.0 & 74.3 & \textbf{50.0} & \textbf{88.2}\\
\midrule
\multirow{3}{*}{0.5}
& SnapKV & \textbf{100.0} & 94.2 & 15.9 & 93.5 & 48.3 & 15.3 & 77.9 & 87.8 & 94.8 & \textbf{72.3} & 85.5 & 44.6 & 37.5 & 66.8\\
& StreamingLLM & 47.2 & 46.0 & 46.4 & 53.2 & 50.0 & \textbf{44.4} & 48.5 & 52.4 & 69.5 & 1.6 & 83.1 & \textbf{75.7} & 35.9 & 50.3\\
& LagKV & \textbf{100.0} & \textbf{98.8} & \textbf{88.4} & \textbf{98.7} & \textbf{81.7} & 13.9 & \textbf{97.9} & \textbf{98.5} & \textbf{98.7} & 65.7 & \textbf{86.3} & 66.2 & \textbf{45.3} & \textbf{80.0}\\
\midrule
\multirow{3}{*}{0.75}
& SnapKV & 93.4 & 79.3 & 4.3 & 52.0 & 26.7 & 1.4 & 33.5 & 35.7 & 83.0 & \textbf{17.1} & 77.2 & 28.4 & 26.6 & 43.0\\
& StreamingLLM & 28.6 & 21.8 & 20.3 & 33.8 & 26.7 & \textbf{23.6} & 23.2 & 27.1 & 43.6 & 0.9 & \textbf{80.4} & 33.8 & 29.7 & 30.3\\
& LagKV & \textbf{100.0} & \textbf{98.8} & \textbf{46.4} & \textbf{90.9} & \textbf{33.3} & 1.4 & \textbf{86.2} & \textbf{92.4} & \textbf{96.1} & 10.9 & 73.3 & \textbf{46.0} & \textbf{42.2} & \textbf{62.9}\\
\midrule
\multirow{3}{*}{0.875}
& SnapKV & 85.7 & 43.7 & 4.3 & 26.0 & 15.0 & 1.4 & 17.1 & 14.3 & 61.3 & \textbf{1.9} & 66.3 & 18.9 & 26.6 & 29.4\\
& StreamingLLM & 12.1 & 13.8 & \textbf{11.6} & 26.0 & \textbf{16.7} & \textbf{15.3} & 12.3 & 13.4 & 20.7 & 0.9 & \textbf{75.3} & \textbf{29.7} & 29.7 & 21.3\\
& LagKV & \textbf{95.6} & \textbf{77.0} & 5.8 & \textbf{75.3} & 8.3 & 1.4 & \textbf{70.6} & \textbf{80.8} & \textbf{89.8} & 1.5 & 62.0 & \textbf{29.7} & \textbf{37.5} & \textbf{48.9}\\
\bottomrule

\end{tabular}
}
\end{table*}

\begin{table*}[ht!]
\centering
\caption{RULER-16K Results of Qwen2.5-7B-Instruct}
\label{ruler_qwen}
\resizebox{\textwidth}{!}{
\begin{tabular}{c|c|ccccccccccccc|c}
\toprule
\textbf{Comp. Ratio}   & Method & SK1 & SK2 & SK3 & MK1 & MK2 & MK3 & MV & MQ & VT & CWE & FWE & QA1 & QA2 & AVERAGE \\
\midrule
\multirow{1}{*}{0.0}
& FullKV & 100.0 & 100.0 & 100.0 & 99.2 & 99.1 & 94.2 & 94.3 & 100.0 & 99.0 & 79.9 & 93.2 & 72.6 & 48.2 & 90.8\\
\midrule
\multirow{3}{*}{0.25}
& SnapKV & 88.2 & 90.6 & 4.5 & 44.5 & 57.1 & 50.0 & 39.8 & 44.9 & 92.2 & \textbf{80.1} & \textbf{92.8} & 62.9 & 42.0 & 60.7\\
& StreamingLLM & 76.3 & 72.5 & 75.9 & 78.9 & 79.5 & \textbf{67.5} & 71.1 & 74.4 & 76.8 & 74.1 & 89.3 & \textbf{69.3} & 36.6 & 72.5\\
& LagKV & \textbf{100.0} & \textbf{99.3} & \textbf{86.6} & \textbf{98.4} & \textbf{88.4} & 24.2 & \textbf{93.9} & \textbf{99.4} & \textbf{99.0} & 79.3 & 92.1 & 66.1 & \textbf{45.5} & \textbf{82.5}\\
\midrule
\multirow{3}{*}{0.5}
& SnapKV & 86.8 & 64.5 & 3.6 & 24.2 & 28.6 & 9.2 & 21.9 & 23.0 & 90.8 & \textbf{77.8} & \textbf{92.3} & 40.3 & 35.7 & 46.1\\
& StreamingLLM & 50.7 & 42.8 & \textbf{52.7} & 52.3 & 45.5 & \textbf{45.8} & 48.8 & 51.2 & 61.5 & 72.3 & 88.3 & \textbf{72.6} & 33.0 & 55.2\\
& LagKV & \textbf{100.0} & \textbf{97.8} & 48.2 & \textbf{98.4} & \textbf{54.5} & 3.3 & \textbf{93.9} & \textbf{95.3} & \textbf{98.6} & 74.4 & 89.5 & 58.1 & \textbf{42.9} & \textbf{73.5}\\
\midrule
\multirow{3}{*}{0.75}
& SnapKV & 82.2 & 18.8 & 3.6 & 13.3 & 10.7 & 4.2 & 13.2 & 12.2 & 79.7 & 64.4 & \textbf{89.3} & 27.4 & 27.7 & 34.4\\
& StreamingLLM & 25.0 & 20.3 & \textbf{22.3} & 28.9 & \textbf{25.9} & \textbf{20.0} & 24.4 & 25.8 & 35.8 & \textbf{66.9} & 82.3 & 32.3 & 24.1 & 33.4\\
& LagKV & \textbf{99.3} & \textbf{87.7} & 8.9 & \textbf{85.2} & 6.2 & 0.8 & \textbf{86.2} & \textbf{83.5} & \textbf{95.6} & 43.8 & 69.5 & \textbf{39.5} & \textbf{29.5} & \textbf{56.6}\\
\midrule
\multirow{3}{*}{0.875}
& SnapKV & 69.7 & 8.0 & 3.6 & 14.1 & 6.2 & 0.8 & 11.6 & 11.4 & 55.1 & 47.7 & \textbf{79.2} & 19.4 & \textbf{22.3} & 26.9\\
& StreamingLLM & 11.2 & 13.8 & \textbf{14.3} & 18.8 & \textbf{15.2} & \textbf{12.5} & 13.4 & 14.0 & 20.7 & \textbf{56.1} & 78.1 & 21.0 & 18.8 & 23.7\\
& LagKV & \textbf{99.3} & \textbf{62.3} & 3.6 & \textbf{56.2} & 0.0 & 0.8 & \textbf{60.2} & \textbf{50.6} & \textbf{93.0} & 18.8 & 55.2 & \textbf{29.0} & \textbf{22.3} & \textbf{42.4}\\
\bottomrule

\end{tabular}
}
\end{table*}

\section{Ablations}

In this section, we fix the sink size to $S=16$ and vary the lag size $L$ and retention ratio $r$. The values of $L$ will be $L=128$, $512$ and $1024$. The values of $r$ will be $2 \times$, $4 \times$, $6 \times$ and $8 \times$ which correspond to $r=0.5$, $0.25$, $0.167$, and $0.125$ respectively. Aslo, we will alter the prefilling method to prove the stability of our approach and scoring method for the validity of lag information. 

{\bf Datasets.} We use the facility in~\cite{yuan_liu_zhong_2024_kvcache_comp_benchmnark} to extensively test our method.  It contains two benchmarks: LongBench~\cite{bai2023longbench} and Needle-in-a-HaystackTest with Passkey-Retrieval in a Paul Graham Essays background~\cite{needle,mohtashami2023landmarkattentionrandomaccessinfinite}.  We only test the 64-digit passkey retrieval task which is much more challenging. And because we are using a recursive and evicting compression strategy, it's easier to illustrate some insights with the partial match score other than the exact one in their report. Therefore, the default needle score will be the partial score throughout the work unless otherwise specified. The main result of this ablation is Table~\ref{tab:main_table}. 

{\bf Prefill stage.} By default, like many other compression methods, compression begins after prefill completes for each layer. This is an efficient and accurate approach—preserving both the KV cache values and the first generated token (FGT) while reducing KV cache size. However, since we lack a reliable benchmark for long-context and long-generation scenarios, we will extend the passkey retrieval task by enabling chunk-by-chunk compression during prefill. This will help us evaluate how compression impacts long-generation performance, especially the FGT. Also, this chunked prefilling method will be useful for extreme long context processing.

\begin{table*}[ht!]
\centering
\caption{Performance of LagKV.}

\resizebox{\textwidth}{!}{
\begin{tabular}{c|c|cccccc|cc}
\toprule
\textbf{Model}      & \textbf{Method} & Single. QA & Multi. QA & Summ. & Few-shot & Synthetic & Code & \textbf{LB Avg.} & \textbf{Needle} \\
\midrule
\multirow{13}{*}{\rotatebox[origin=c]{90}{Llama-3.1-8B-Instruct}} 
& FullKV & 40.71 & 37.90 & 28.29 & 68.49 & 68.00 & 58.70 & 47.44 & 99.44 \\
\cline{2-10}
& L=1024,r=2x & 39.42 & 37.12 & 27.38 & 67.71 & 68.50 & 58.83 & 46.74 & 99.27 \\
& L=1024,r=4x & 37.06 & 36.77 & 26.79 & 66.96 & 63.50 & 58.42 & 45.54 & 96.57 \\
& L=1024,r=6x & 35.74 & 36.08 & 26.33 & 66.33 & 60.50 & 57.91 & 44.65 & 91.77 \\
& L=1024,r=8x & 35.49 & 35.99 & 25.90 & 65.21 & 61.00 & 57.95 & 44.31 & 86.26 \\
\cline{2-10}
& L=512,r=2x & 39.43 & 37.45 & 27.35 & 67.82 & 67.50 & 58.66 & 46.73 & 97.02 \\
& L=512,r=4x & 37.39 & 36.27 & 26.19 & 66.56 & 62.50 & 57.86 & 45.16 & 85.73 \\
& L=512,r=6x & 34.95 & 35.62 & 25.52 & 65.87 & 59.50 & 58.14 & 44.11 & 75.67 \\
& L=512,r=8x & 34.03 & 36.12 & 25.25 & 64.94 & 56.00 & 57.50 & 43.47 & 68.25 \\
\cline{2-10}
& L=128,r=2x & 38.56 & 36.80 & 27.20 & 67.64 & 68.00 & 59.27 & 46.48 & 92.76 \\
& L=128,r=4x & 36.66 & 36.58 & 25.62 & 66.78 & 66.50 & 57.90 & 45.28 & 73.41 \\
& L=128,r=6x & 34.57 & 35.41 & 24.59 & 63.59 & 64.00 & 56.97 & 43.49 & 38.48 \\
& L=128,r=8x & 33.78 & 34.60 & 23.91 & 62.21 & 61.50 & 55.68 & 42.42 & 25.01 \\
\midrule
\multirow{13}{*}{\rotatebox[origin=c]{90}{Qwen-2.5-7B-Instruct}} 
& FullKV & 41.62 & 45.00 & 26.41 & 68.91 & 100.00 & 63.60 & 51.53 & 100.00 \\
\cline{2-10}
& L=1024,r=2x & 39.80 & 42.85 & 26.11 & 67.66 & 99.50 & 63.12 & 50.33 & 99.75 \\
& L=1024,r=4x & 36.92 & 40.39 & 24.81 & 65.91 & 95.00 & 61.60 & 48.15 & 96.98 \\
& L=1024,r=6x & 35.77 & 39.74 & 24.68 & 65.28 & 93.50 & 61.45 & 47.52 & 77.47 \\
& L=1024,r=8x & 34.60 & 39.10 & 24.18 & 64.74 & 90.50 & 61.30 & 46.73 & 66.88 \\
\cline{2-10}
& L=512,r=2x & 38.72 & 42.79 & 25.91 & 67.98 & 98.50 & 62.00 & 49.91 & 97.07 \\
& L=512,r=4x & 35.42 & 39.12 & 24.49 & 64.59 & 94.00 & 60.16 & 47.01 & 75.89 \\
& L=512,r=6x & 34.00 & 38.04 & 23.72 & 64.31 & 87.50 & 58.80 & 45.69 & 42.70 \\
& L=512,r=8x & 32.14 & 37.83 & 23.11 & 63.48 & 82.50 & 58.71 & 44.64 & 30.00 \\
\cline{2-10}
& L=128,r=2x & 38.67 & 42.49 & 25.69 & 67.75 & 99.00 & 60.64 & 49.61 & 65.93 \\
& L=128,r=4x & 34.47 & 39.78 & 24.07 & 65.13 & 96.00 & 58.67 & 46.91 & 20.83 \\
& L=128,r=6x & 32.83 & 38.15 & 22.95 & 62.23 & 90.50 & 56.25 & 44.76 & 16.18 \\
& L=128,r=8x & 32.47 & 37.10 & 22.20 & 60.24 & 88.50 & 56.10 & 43.78 & 15.07 \\
\bottomrule

\end{tabular}
}
\label{tab:main_table}
\end{table*}

\subsection{LongBench}
For the LongBench dataset, the 
 method performs very well across different ratios and lag sizes. When $L=1024,r=8 \times$, the method still retains approximate $90\%$ of the baseline performance. Since the compression ratio will increase when $L$ decreases, the worse case is $L=128,r=8 \times$ for both models but the method maintains at least $85\%$ of the baseline performance.

\subsection{Passkey Retrieval}
The 64-digit passkey retrieval task  is a challenging one for most token eviction strategies. As discussed in \cite{yuan_liu_zhong_2024_kvcache_comp_benchmnark}, the most successful eviction strategy $H_2O$ \cite{H2o} performs well in 7-digit task (scoring $100\%$ for all compression ratios) but degrades a lot in the 64-digit one (for $4 \times$ in Llama-3, exact match score is $35\%$ and partial match score is $70.8\%$). It happens because the strategy applies its compression after the prefill is done which means the FGT is not affected by the compression and the 7-digit passkey usually takes only 2 or 3 tokens. When the passkey size increases to 64, much more generated tokens are impacted by the compression. Many token-evict algorithms are struggling to maintain their performance in this case. In contrast, our method performs very well when the product of $r$ and $L$ is sufficient large enough (for $L=1024,r=4 \times$ in Llama model, exact math score is $89\%$ and partial match score is $96.57\%$).

\begin{figure}[h]
\centering
\begin{minipage}{\linewidth}
\centerline{\includegraphics[width=\textwidth]{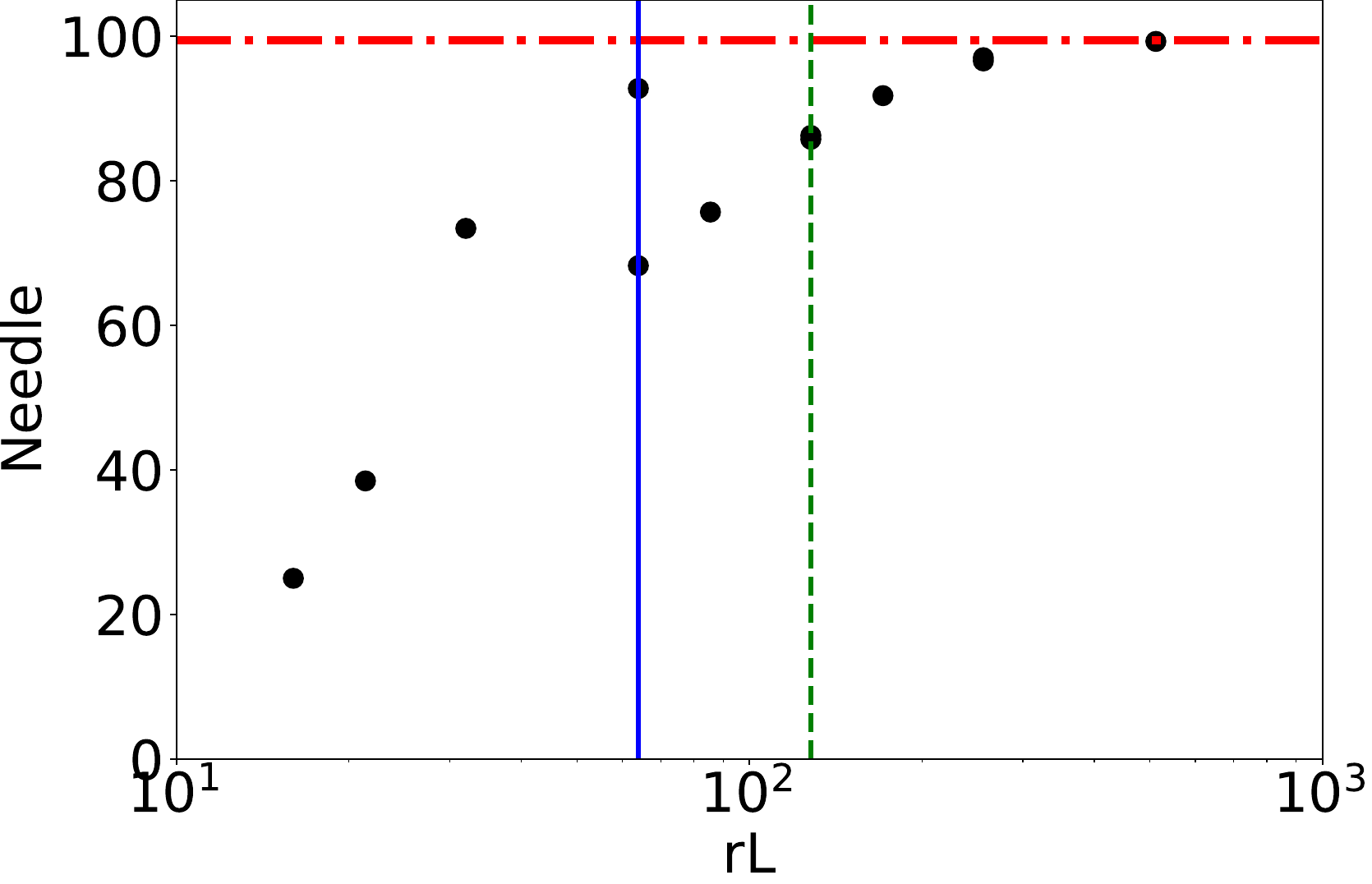}}
\centerline{(a) Llama-3.1-8B}
\end{minipage}
\begin{minipage}{\linewidth}
\centerline{\includegraphics[width=\textwidth]{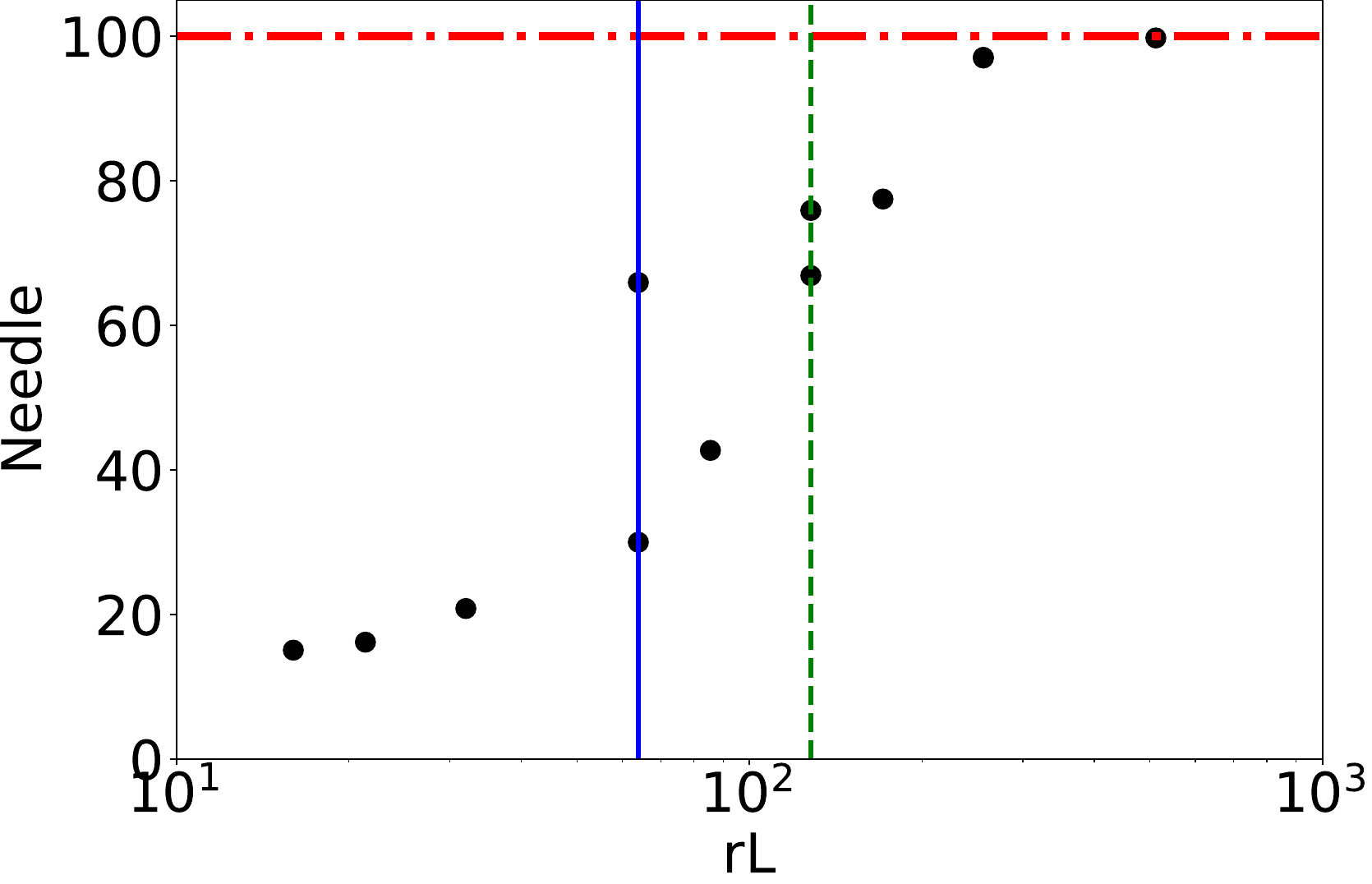}}
\centerline{(b) Qwen-2.5-7B}
\end{minipage}
\caption{The needle score vs different setups of rL. The horizontal dash-dot line is the baseline for each model. The x-axis is in log scale. We put two vertical lines $rL=64$ (solid blue) and $rL=128$ (dash green) for guidelines.}
\label{fig:needle_vs_rL}
\end{figure}

Our recursive compression strategy will not perform well for the setups with small $rL$ due to the fact that when the recursive window size is compressed to be close to or less than the length of the queried content, it's highly possible that only a small portion of the wanted information will be kept. In the task of 64-digit passkey retrieval, because digits usually require more tokens to be represented than the same length words, the number of expected tokens is much larger than the similar tasks in LongBench sub tasks like Document QA, that leads to its results are more sensitive to small $rL$. As shown in Fig.~\ref{fig:needle_vs_rL}, the Qwen model which uses one token for one digit degenerates faster than the Llama model which represents three digits by one token with smaller $rL$. It hints us that we must choose the compression ratio and the lag size carefully in considering the length of the expected content and the tokenizer of the LLM. ~\ref{apdx:needledetails} shows all the details of the needle results.

\subsection{Chunk-by-Chunk Compression in Prefill Stage}

\begin{figure}[h]
\centering
\begin{minipage}{\linewidth}
\centerline{\includegraphics[width=\textwidth]{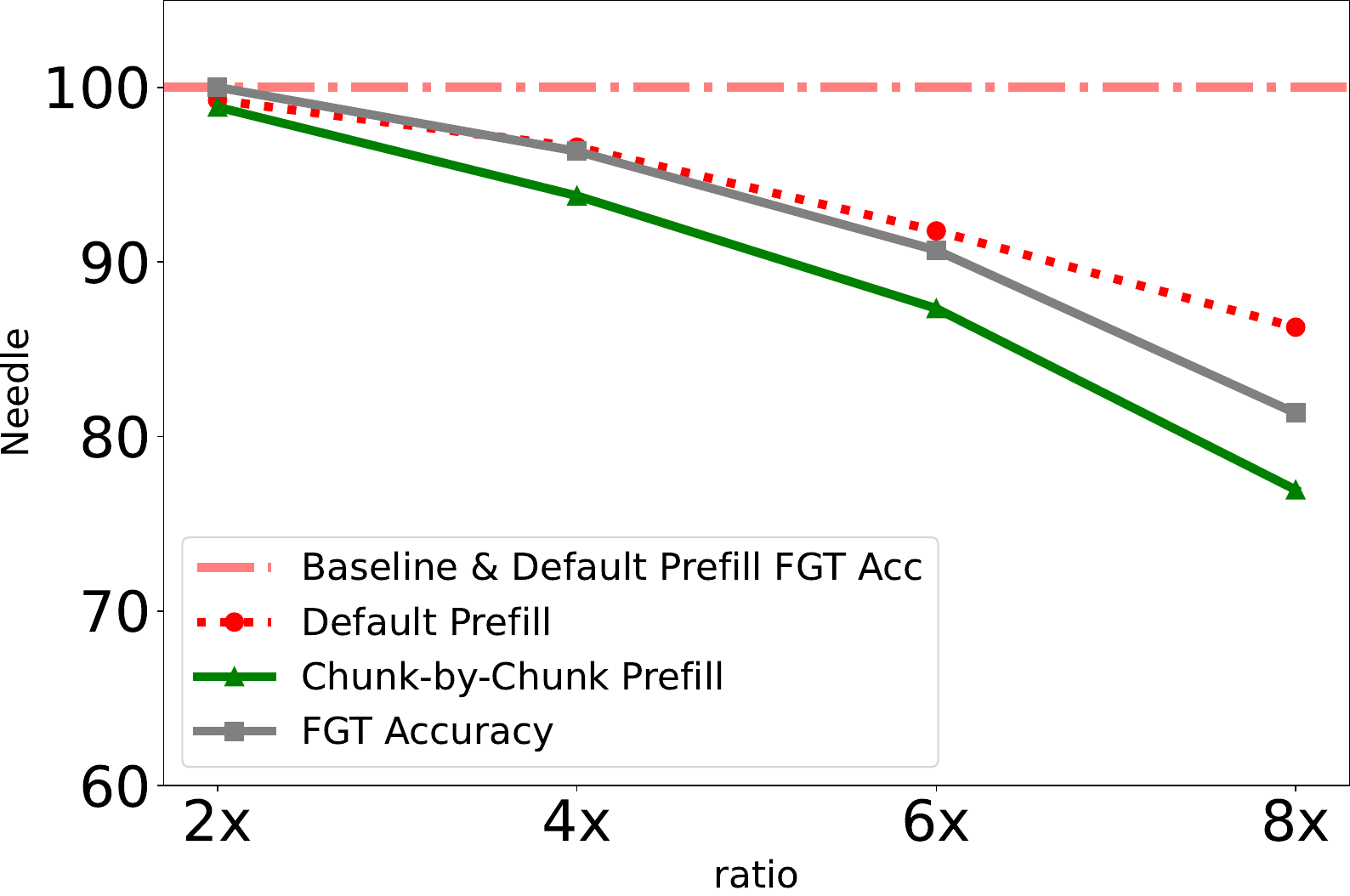}}
\centerline{(a) Llama-3.1-8B}
\end{minipage}
\begin{minipage}{\linewidth}
\centerline{\includegraphics[width=\textwidth]{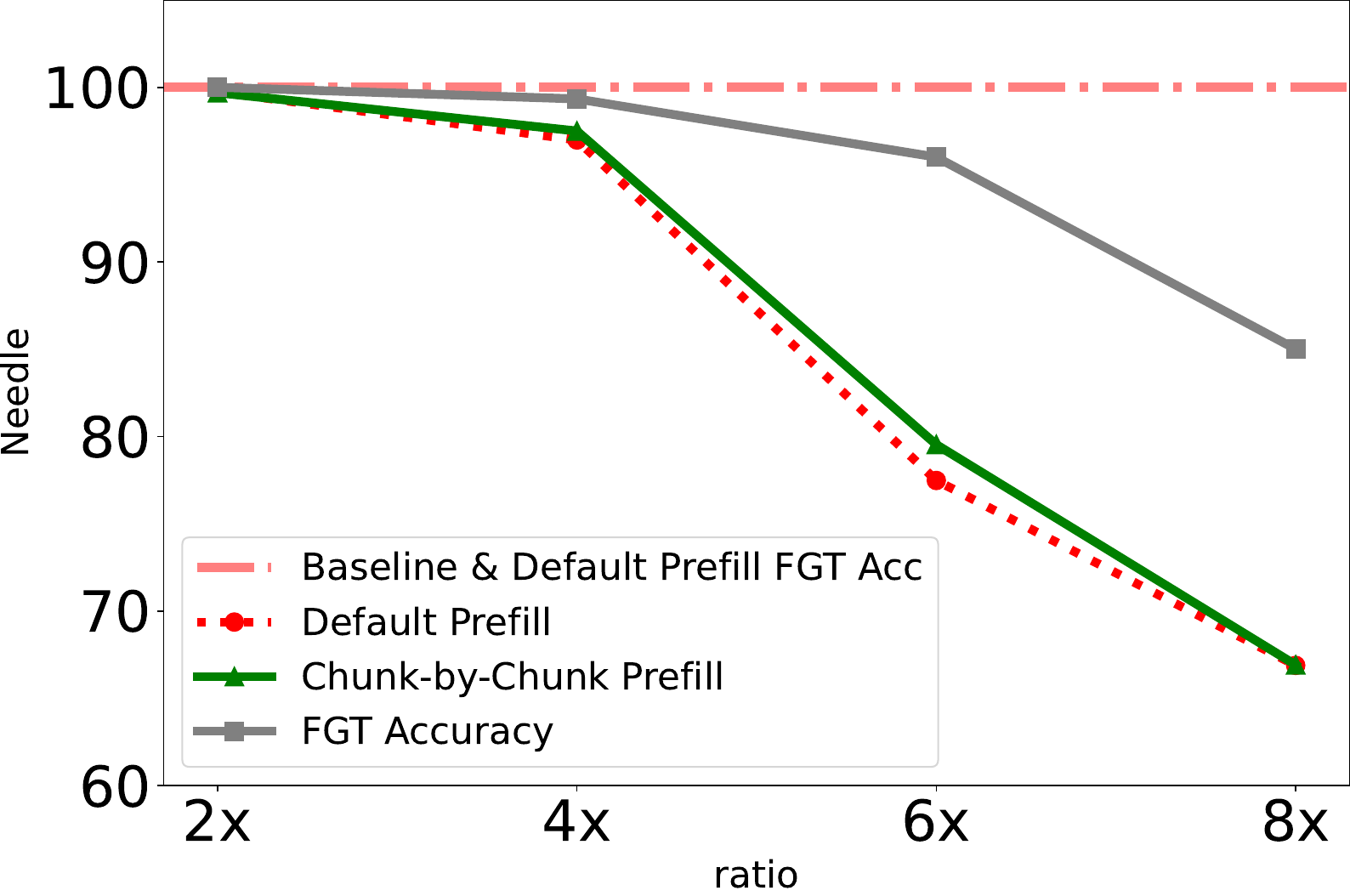}}
\centerline{(b) Qwen-2.5-7B}
\end{minipage}
\caption{The needle score and FGT accuracy for different prefill methods with $L=1024$ only. The horizontal dash-dot line is the baseline for both needle scores and FGT accuracy since they are 
overlapping.}
\label{fig:prefill_fgt}
\end{figure}

To enable chunk-by-chunk compression during prefill, we have to split the retrieval tokens like our recursive compression for long context with prefilling the first $S+2L$ tokens and then $L$ each time until all input tokens are prefilled. In such a way, the hidden values after the first chunk will be different from default prefill ones since less tokens are seen in the forwarding. Then, the FGT may be different too. With the chunk-by-chunk prefill compression, we calculated the FGT accuracy which is defined as the ratio of FGT same as the default prefill ones and also the overall needle scores, shown in Fig.~\ref{fig:prefill_fgt}.

\begin{figure*}[t]

	\centering
	\begin{minipage}{0.24\textwidth}
		\centerline{\includegraphics[width=\textwidth]{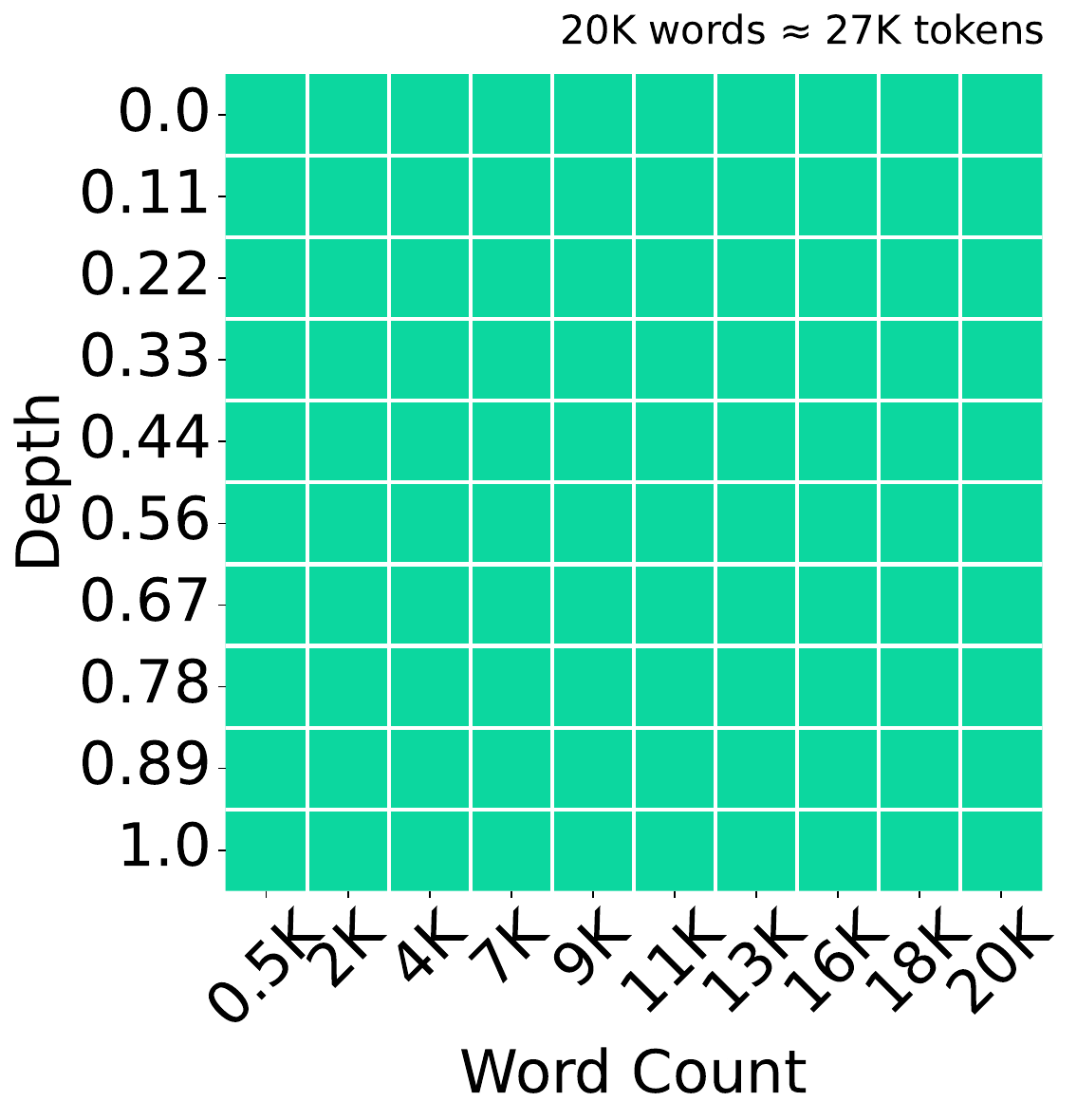}}
		\centerline{$L=1024,r=2 \times$}
	\end{minipage}
	\begin{minipage}{0.24\textwidth}
		\centerline{\includegraphics[width=\textwidth]{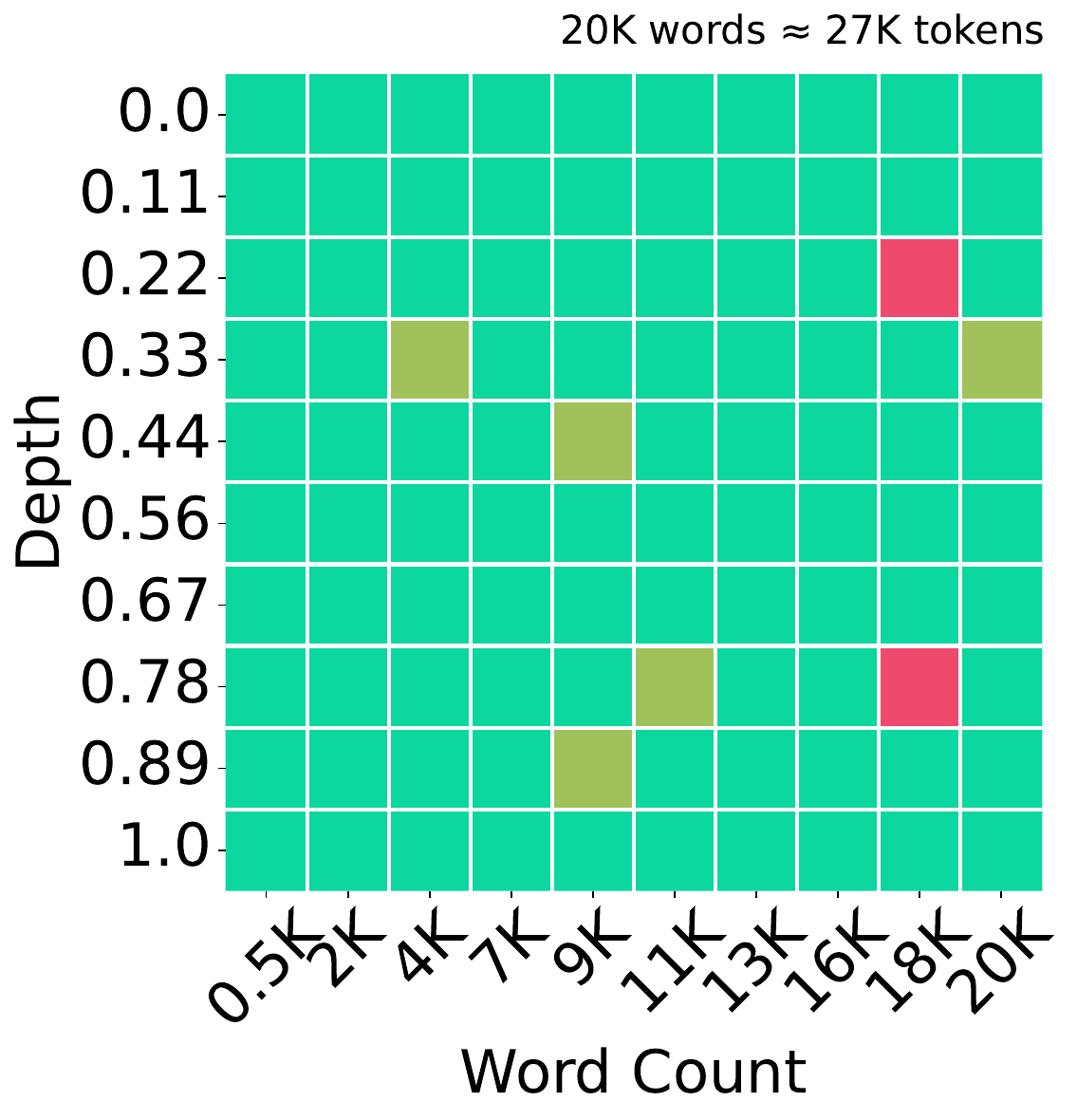}}
		\centerline{$L=1024,r=4 \times$}
	\end{minipage}
\begin{minipage}{0.24\textwidth}
		\centerline{\includegraphics[width=\textwidth]{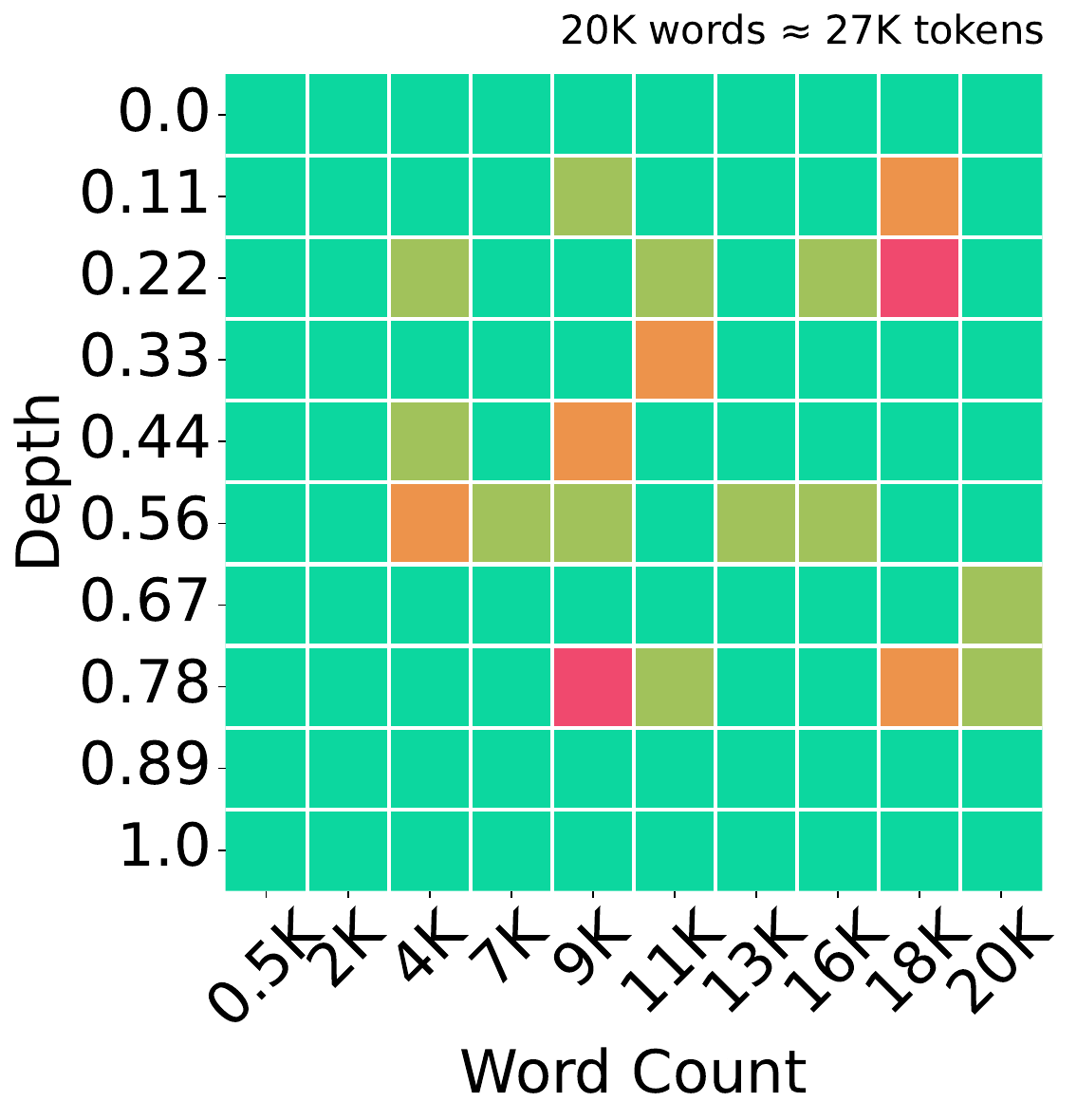}}
		\centerline{$L=1024,r=6 \times$}
	\end{minipage}
	\begin{minipage}{0.24\textwidth}
		\centerline{\includegraphics[width=\textwidth]{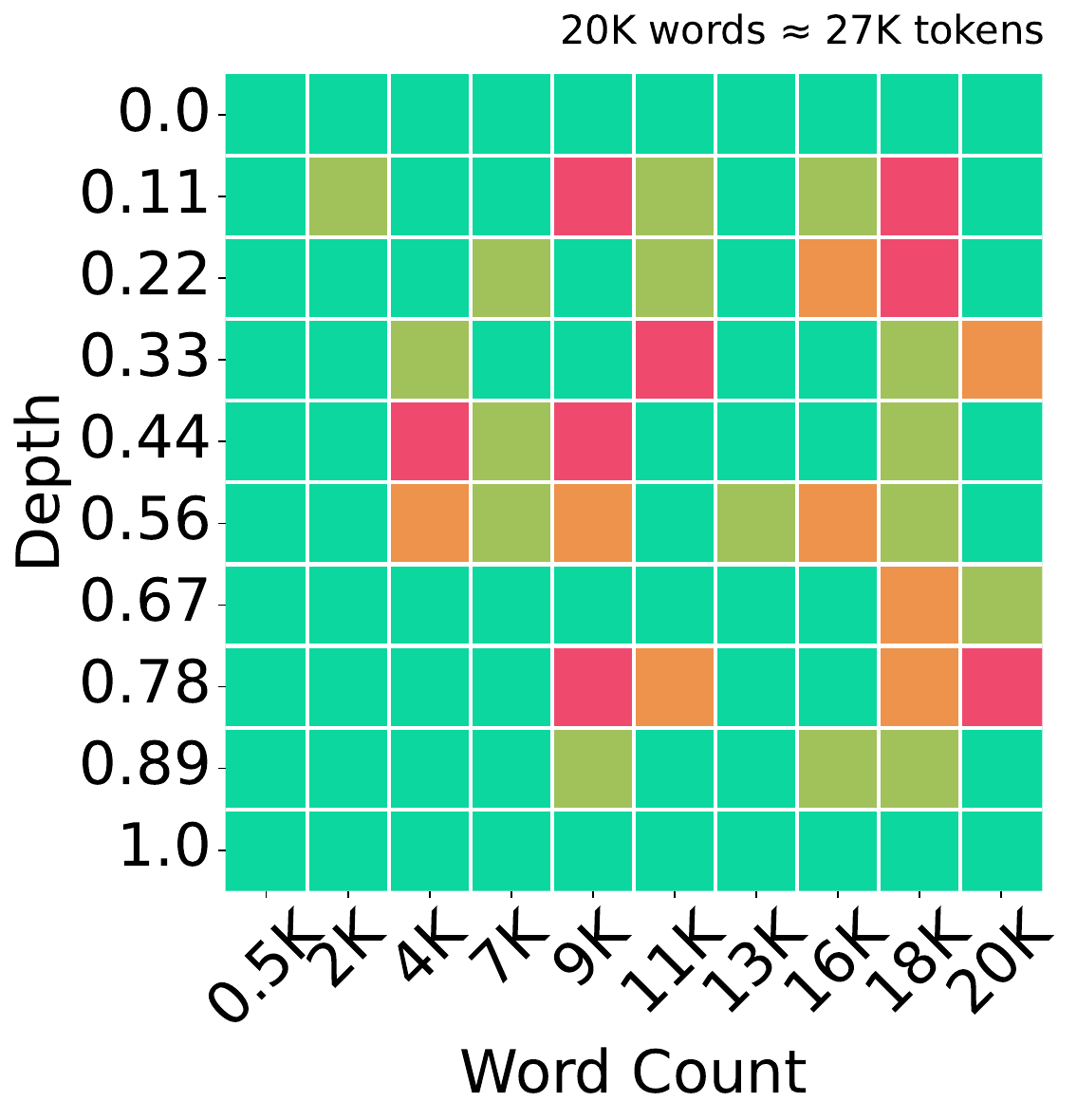}}
		\centerline{$L=1024,r=8 \times$}
	\end{minipage}
	\centerline{(a) Llama-3.1-8B-Instruct}
	\begin{minipage}{0.24\textwidth}
		\centerline{\includegraphics[width=\textwidth]{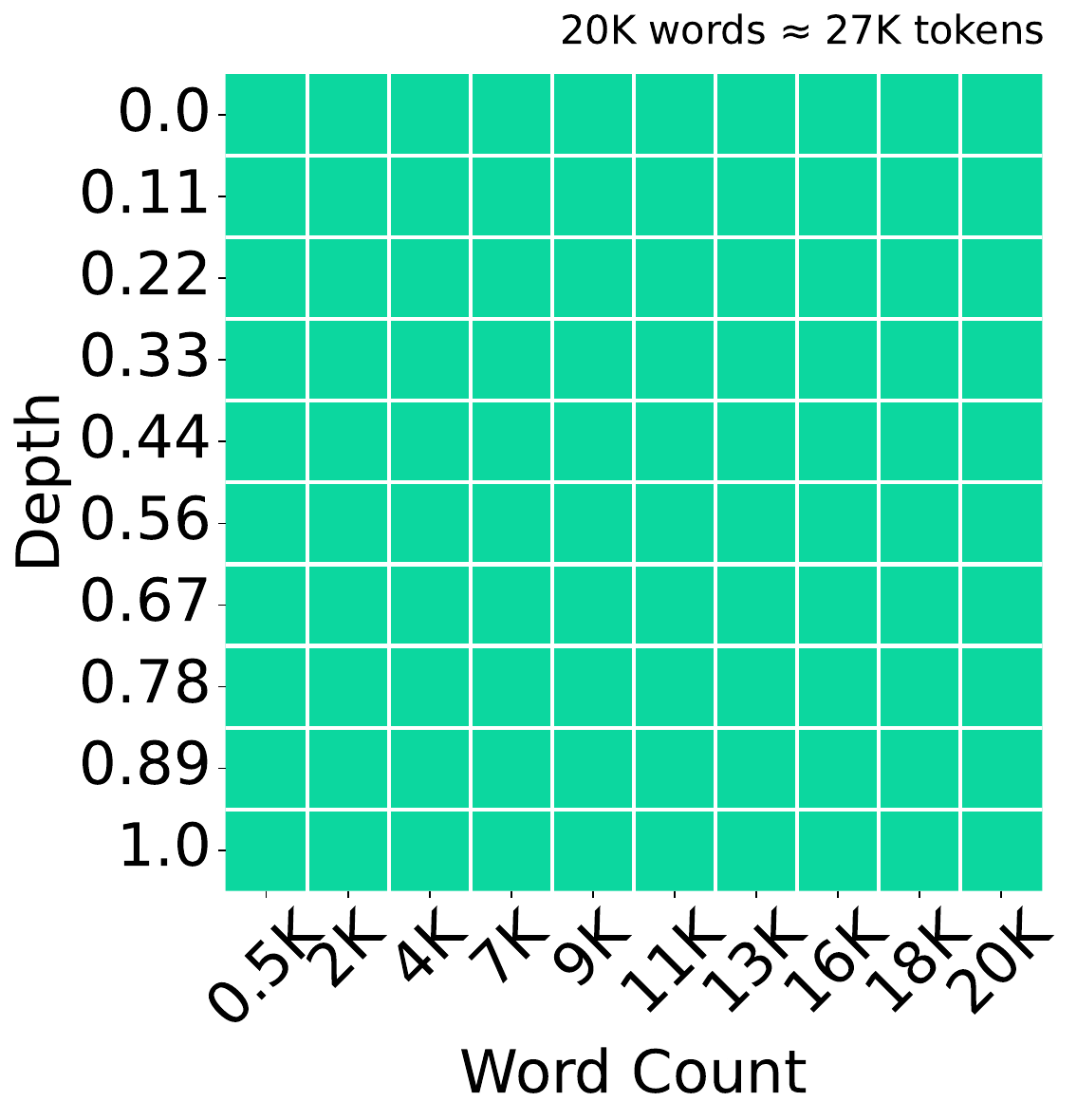}}
		\centerline{$L=1024,r=2 \times$}
	\end{minipage}
	\begin{minipage}{0.24\textwidth}
		\centerline{\includegraphics[width=\textwidth]{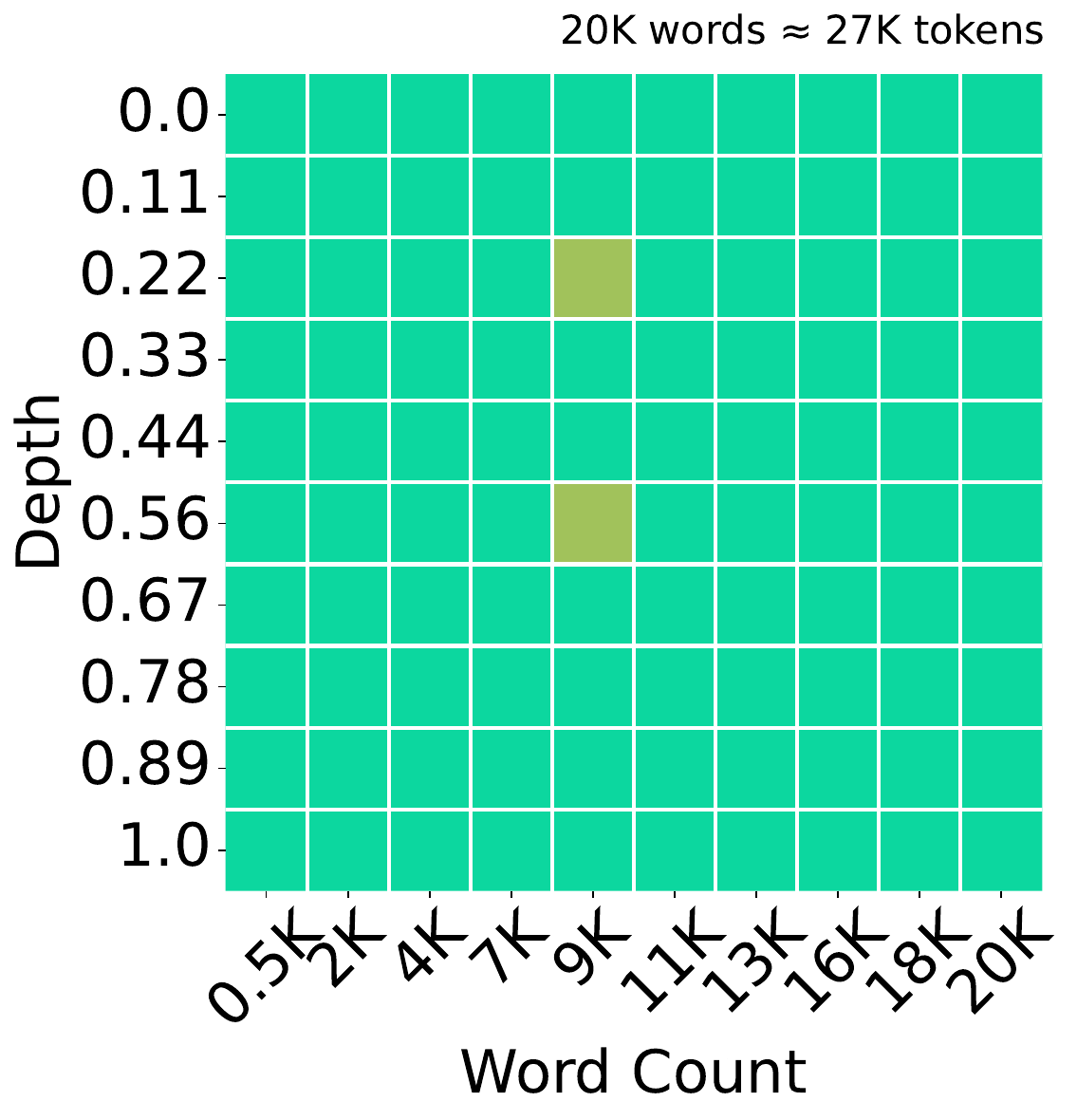}}
		\centerline{$L=1024,r=4 \times$}
	\end{minipage}
\begin{minipage}{0.24\textwidth}
		\centerline{\includegraphics[width=\textwidth]{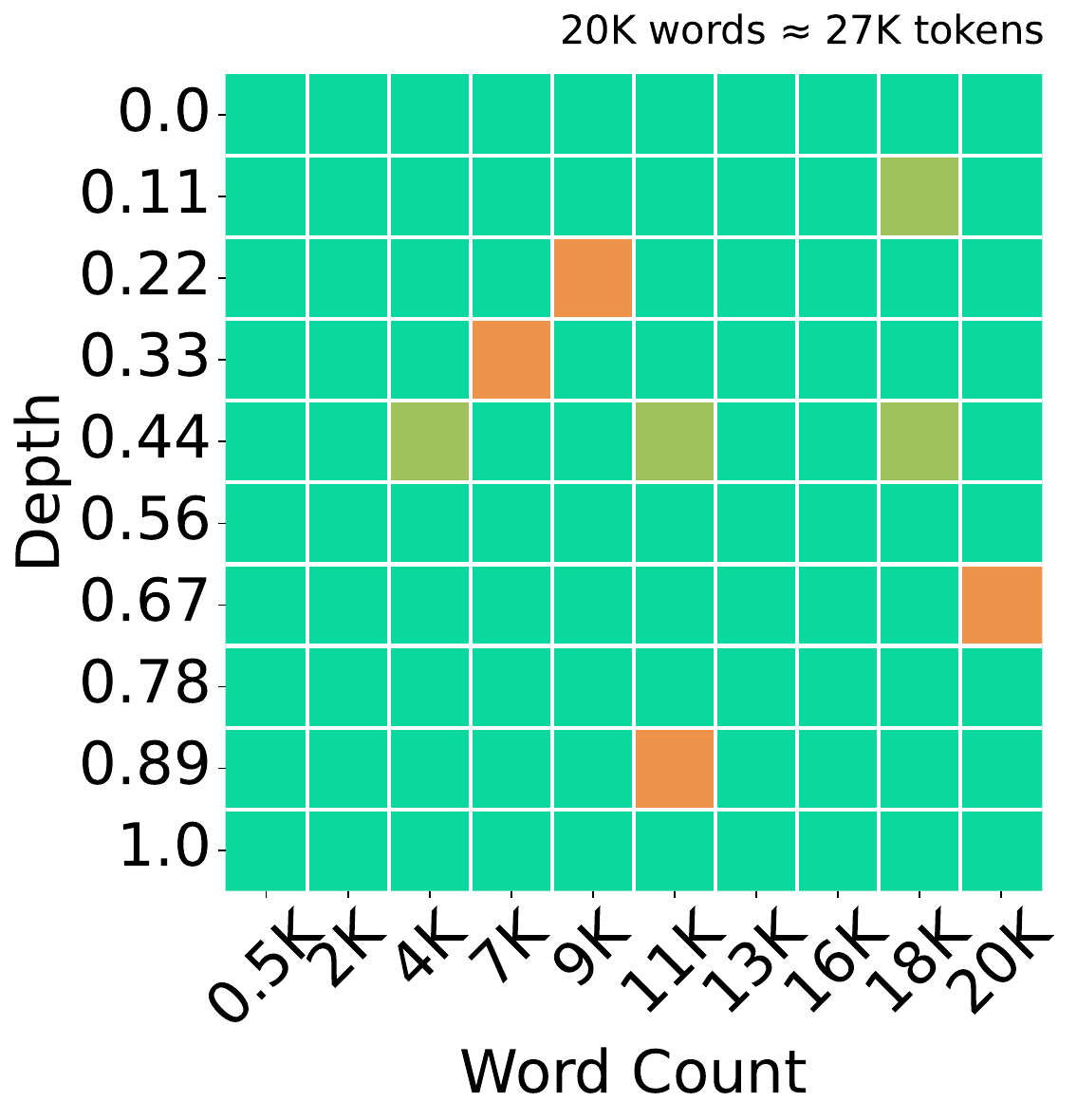}}
		\centerline{$L=1024,r=6 \times$}
	\end{minipage}
	\begin{minipage}{0.24\textwidth}
		\centerline{\includegraphics[width=\textwidth]{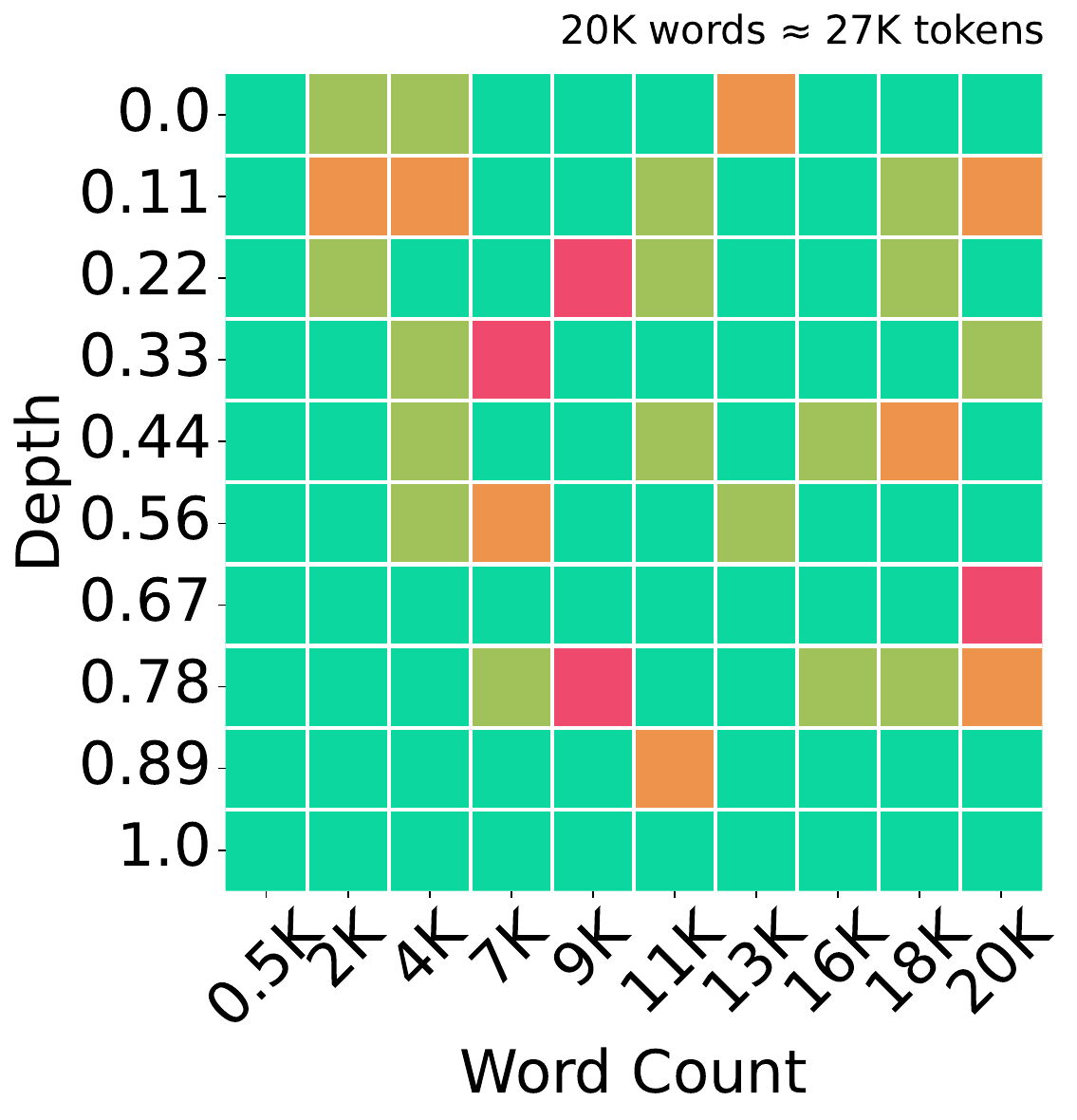}}
		\centerline{$L=1024,r=8 \times$}
	\end{minipage}
	\centerline{(b) Qwen-2.5-7B-Instruct}
	\caption{First Generated Token Accuracy for different setups, sequence lengths and needle depths with chunked prefill. It tests three trials on each depth.}
	\label{fig:FGTHeatmap}
\end{figure*}

The chunked prefill definitely diminishes the FGT accuracy as it drops from $100\%$ to around $80\%$ for $r=8 \times$ in both models. But we do not see it has a strong dependence on sequence lengths or needle depths in Fig.~\ref{fig:FGTHeatmap}. These confirm that our method is able to retain the major part of the baseline capabilities in the case with long sequence hidden values impacted by the compression. It ensures that LagKV will deliver a good performance in the long generation scenarios.

Meanwhile, we also notice that the FGT accuracy and overall needle scores suffer more degradation in Llama model with chunked prefill. It is mainly because different models exhibit various abilities of stable long generation~\cite{quan2024languagemodelsselflengthengenerate}. 

\subsection{Scoring Methods}
We present two different scoring variants from LagKV. Both of them will only change the scoring methods but keep the attention sink and sliding window unchanged. And we only use the 64-digits passkey retrieval task which can easily distinguish eviction strategies as the detector. Among these tests, we keep $S=16,L=1024$ as constant.

The first one is called LocalKV which only skips using the reference from the next joint chunk tokens but replacing the equation Eq. \ref{eq:min} and \ref{eq:max} by the following equations:
\begin{equation}
min_i^{p,Z} = min_{seq}({Z_i^{p}})
\label{eq:min_var}
\end{equation}
\begin{equation}
max_i^{p,Z}=max_{seq}({Z_i^{p}})
\label{eq:max_var}
\end{equation}
Therefore, the min-max is totally from the local chunk instead of the remote one.

The second one is $L_2$ norm from ~\cite{devoto2024simpleeffective}. We adapt the low key states norm method into the recursive framework by replacing Eq.\ref{eq:scoresum} by:
\begin{equation}
score_{i} = -Norm(K_i)
\end{equation}
As suggested in their work, we skip the compression of the first two layers in this variant too.

The results of the 64-digit passkey retrieval task are present in Fig.~\ref{fig:needle1024variants} and ~\ref{fig:needle1024variantsexact} with partial match scores and exact match scores. As we can see, the LagKV method is always the best one especially in the high compression ratios and the exact match cases. The LocalKV variant performs closely to LagKV at low compression ratios but degrades significantly at higher ones. This behavior stems from the similarity between local and remote max-min statistical values, which aligns with the token-wise locality. 

Since the setup of $L=1024$ will have a chunk that fully covers the passkey when the context is shorter than 2K or the passkey is at $100\%$ depth, the bottom line of the exact match score will be about $27\%$ if the selected tokens did not mess up the output. That means the $L_2$ norm variant shows very limited performance with a constant exact match score $27\%$ for all compression ratios and models. 

\begin{figure}

	\centering
	\begin{minipage}{0.49\textwidth}
		\centerline{\includegraphics[width=\textwidth]{Figures/diagrams/llama\_needle\_1024\_variants.pdf}}
		\centerline{Llama-3.1-8B-Instruct}
	\end{minipage}
	\begin{minipage}{0.49\textwidth}
		\centerline{\includegraphics[width=\textwidth]{Figures/diagrams/qwen\_needle\_1024\_variants.pdf}}
		\centerline{Qwen-2.5-7B-Instruct}
	\end{minipage}
	\caption{The 64-digit Passkey Retrieval {\bf partial} match scores of different variants and compression ratios.}
	\label{fig:needle1024variants}
\end{figure}

\begin{figure}

	\centering
	\begin{minipage}{0.49\textwidth}
		\centerline{\includegraphics[width=\textwidth]{Figures/diagrams/llama\_exact\_needle\_1024\_variants.pdf}}
		\centerline{Llama-3.1-8B-Instruct}
	\end{minipage}
	\begin{minipage}{0.49\textwidth}
		\centerline{\includegraphics[width=\textwidth]{Figures/diagrams/qwen\_exact\_needle\_1024\_variants.pdf}}
		\centerline{Qwen-2.5-7B-Instruct}
	\end{minipage}
	\caption{The 64-digit Passkey Retrieval {\bf exact} match scores of different variants and compression ratios.}
	\label{fig:needle1024variantsexact}
\end{figure}

\section{Related Works}

The $L_2$ Norm-Based KV compression ~\cite{devoto2024simpleeffective} is an existing eviction approach that relies solely on KV information to compress the KV cache. This method computes token scores using the negative norm of key states. In contrast to our derivation from the autoregressive process and the token-wise locality, their method is formed by comparing the attention loss.

FINCH~\cite{corallo2024finchpromptguidedkeyvaluecache} introduces a prompt-guided KV compression method for the prefill stage, employing a chunk-by-chunk approach with instruction tokens appended to each document chunk. This design ensures the computation of attention submatrices between instructions and document chunks, enabling subsequent KV cache filtering. In contrast, our proposed chunked prefilling method operates without instructions, making it compatible with multi-turn queries. In other words, our approach transforms a causal LLM into a compressor capable of condensing long documents into compressed KV sequences, which can later be decompressed under varying instructions without reconstruction.

\section{Conclusion}

In this study, we propose LagKV, an attention-weight-free token eviction method. It achieves comparable performance on long-context tasks while significantly outperforming mainstream eviction strategies in 64-digit passkey retrieval tasks. These results demonstrate that our method maintains robust long-text retrieval capabilities even at high compression ratios.

Unlike existing approaches, LagKV employs a recursive attention-weight-free strategy in both prefill and decode stages to determine token importance for future processing. It is independent from query states and the rest part of the long prompt. Therefore our method offers a novel perspective on LLM mechanisms, shedding light on their inner workings in a fundamentally different way.


\bibliography{LagKV}
\bibliographystyle{acl_natbib}

\appendix

\onecolumn
\newpage
\definecolor{question_color}{RGB}{0,100,0}
\appendix
\section{Appendix}
\subsection{Detail Rresults of Passkey Retrieval}
\label{apdx:needledetails}
Here, we present all the Needle-in-a-Haystack results with 64-digit Passkey Retrieval for different setups. The partial matching results are in Fig.\ref{fig:llamaneedlespartial} and \ref{fig:qwenneedlespartial} while Fig.\ref{fig:llamaneedlesexact} and \ref{fig:qwenneedlesexact} with exact matching. Overall accuracies are noted within parentheses on the top-right corner of each sub graph.

\begin{figure*}

	\centering
	\begin{minipage}{0.24\textwidth}
		\centerline{\includegraphics[width=\textwidth]{Figures/diagrams/lagkv\_test/paulgraham\_passkey/20480words\_10x10x3\_64digits/lag\_kv\_partial\_match/llama3.1-8b-instruct/lb\_ratio\_\_0.5rb.pdf}}
		\centerline{$L=1024,r=2 \times$}
	\end{minipage}
	\begin{minipage}{0.24\textwidth}
		\centerline{\includegraphics[width=\textwidth]{Figures/diagrams/lagkv\_test/paulgraham\_passkey/20480words\_10x10x3\_64digits/lag\_kv\_partial\_match/llama3.1-8b-instruct/lb\_ratio\_\_0.25rb.pdf}}
		\centerline{$L=1024,r=4 \times$}
	\end{minipage}
\begin{minipage}{0.24\textwidth}
		\centerline{\includegraphics[width=\textwidth]{Figures/diagrams/lagkv\_test/paulgraham\_passkey/20480words\_10x10x3\_64digits/lag\_kv\_partial\_match/llama3.1-8b-instruct/lb\_ratio\_\_0.1667rb.pdf}}
		\centerline{$L=1024,r=6 \times$}
	\end{minipage}
	\begin{minipage}{0.24\textwidth}
		\centerline{\includegraphics[width=\textwidth]{Figures/diagrams/lagkv\_test/paulgraham\_passkey/20480words\_10x10x3\_64digits/lag\_kv\_partial\_match/llama3.1-8b-instruct/lb\_ratio\_\_0.125rb.pdf}}
		\centerline{$L=1024,r=8 \times$}
	\end{minipage}

	\begin{minipage}{0.24\textwidth}
		\centerline{\includegraphics[width=\textwidth]{Figures/diagrams/lagkv\_test/paulgraham\_passkey/20480words\_10x10x3\_64digits/lag\_kv\_partial\_match/llama3.1-8b-instruct/lb\_ratio\_\_0.5,\_lag\_size\_\_512rb.pdf}}
		\centerline{$L=512,r=2 \times$}
	\end{minipage}
	\begin{minipage}{0.24\textwidth}
		\centerline{\includegraphics[width=\textwidth]{Figures/diagrams/lagkv\_test/paulgraham\_passkey/20480words\_10x10x3\_64digits/lag\_kv\_partial\_match/llama3.1-8b-instruct/lb\_ratio\_\_0.25,\_lag\_size\_\_512rb.pdf}}
		\centerline{$L=512,r=4 \times$}
	\end{minipage}
\begin{minipage}{0.24\textwidth}
		\centerline{\includegraphics[width=\textwidth]{Figures/diagrams/lagkv\_test/paulgraham\_passkey/20480words\_10x10x3\_64digits/lag\_kv\_partial\_match/llama3.1-8b-instruct/lb\_ratio\_\_0.1667,\_lag\_size\_\_512rb.pdf}}
		\centerline{$L=512,r=6 \times$}
	\end{minipage}
	\begin{minipage}{0.24\textwidth}
		\centerline{\includegraphics[width=\textwidth]{Figures/diagrams/lagkv\_test/paulgraham\_passkey/20480words\_10x10x3\_64digits/lag\_kv\_partial\_match/llama3.1-8b-instruct/lb\_ratio\_\_0.125,\_lag\_size\_\_512rb.pdf}}
		\centerline{$L=512,r=8 \times$}
	\end{minipage}

	\begin{minipage}{0.24\textwidth}
		\centerline{\includegraphics[width=\textwidth]{Figures/diagrams/lagkv\_test/paulgraham\_passkey/20480words\_10x10x3\_64digits/lag\_kv\_partial\_match/llama3.1-8b-instruct/lb\_ratio\_\_0.5,\_lag\_size\_\_128rb.pdf}}
		\centerline{$L=128,r=2 \times$}
	\end{minipage}
	\begin{minipage}{0.24\textwidth}
		\centerline{\includegraphics[width=\textwidth]{Figures/diagrams/lagkv\_test/paulgraham\_passkey/20480words\_10x10x3\_64digits/lag\_kv\_partial\_match/llama3.1-8b-instruct/lb\_ratio\_\_0.25,\_lag\_size\_\_128rb.pdf}}
		\centerline{$L=128,r=4 \times$}
	\end{minipage}
\begin{minipage}{0.24\textwidth}
		\centerline{\includegraphics[width=\textwidth]{Figures/diagrams/lagkv\_test/paulgraham\_passkey/20480words\_10x10x3\_64digits/lag\_kv\_partial\_match/llama3.1-8b-instruct/lb\_ratio\_\_0.1667,\_lag\_size\_\_128rb.pdf}}
		\centerline{$L=128,r=6 \times$}
	\end{minipage}
	\begin{minipage}{0.24\textwidth}
		\centerline{\includegraphics[width=\textwidth]{Figures/diagrams/lagkv\_test/paulgraham\_passkey/20480words\_10x10x3\_64digits/lag\_kv\_partial\_match/llama3.1-8b-instruct/lb\_ratio\_\_0.125,\_lag\_size\_\_128rb.pdf}}
		\centerline{$L=128,r=8 \times$}
	\end{minipage}

	\caption{The  64-digit Passkey Retrieval of Llama-3.1-8B-Instruct for different setups with {\bf partial} matching.}
	\label{fig:llamaneedlespartial}
\end{figure*}

\begin{figure*}

	\centering
	\begin{minipage}{0.24\textwidth}
		\centerline{\includegraphics[width=\textwidth]{Figures/diagrams/lagkv\_test/paulgraham\_passkey/20480words\_10x10x3\_64digits/lag\_kv\_partial\_match/qwen2.5-7b-instruct/lb\_ratio\_\_0.5rb.pdf}}
		\centerline{$L=1024,r=2 \times$}
	\end{minipage}
	\begin{minipage}{0.24\textwidth}
		\centerline{\includegraphics[width=\textwidth]{Figures/diagrams/lagkv\_test/paulgraham\_passkey/20480words\_10x10x3\_64digits/lag\_kv\_partial\_match/qwen2.5-7b-instruct/lb\_ratio\_\_0.25rb.pdf}}
		\centerline{$L=1024,r=4 \times$}
	\end{minipage}
\begin{minipage}{0.24\textwidth}
		\centerline{\includegraphics[width=\textwidth]{Figures/diagrams/lagkv\_test/paulgraham\_passkey/20480words\_10x10x3\_64digits/lag\_kv\_partial\_match/qwen2.5-7b-instruct/lb\_ratio\_\_0.1667rb.pdf}}
		\centerline{$L=1024,r=6 \times$}
	\end{minipage}
	\begin{minipage}{0.24\textwidth}
		\centerline{\includegraphics[width=\textwidth]{Figures/diagrams/lagkv\_test/paulgraham\_passkey/20480words\_10x10x3\_64digits/lag\_kv\_partial\_match/qwen2.5-7b-instruct/lb\_ratio\_\_0.125rb.pdf}}
		\centerline{$L=1024,r=8 \times$}
	\end{minipage}

	\begin{minipage}{0.24\textwidth}
		\centerline{\includegraphics[width=\textwidth]{Figures/diagrams/lagkv\_test/paulgraham\_passkey/20480words\_10x10x3\_64digits/lag\_kv\_partial\_match/qwen2.5-7b-instruct/lb\_ratio\_\_0.5,\_lag\_size\_\_512rb.pdf}}
		\centerline{$L=512,r=2 \times$}
	\end{minipage}
	\begin{minipage}{0.24\textwidth}
		\centerline{\includegraphics[width=\textwidth]{Figures/diagrams/lagkv\_test/paulgraham\_passkey/20480words\_10x10x3\_64digits/lag\_kv\_partial\_match/qwen2.5-7b-instruct/lb\_ratio\_\_0.25,\_lag\_size\_\_512rb.pdf}}
		\centerline{$L=512,r=4 \times$}
	\end{minipage}
\begin{minipage}{0.24\textwidth}
		\centerline{\includegraphics[width=\textwidth]{Figures/diagrams/lagkv\_test/paulgraham\_passkey/20480words\_10x10x3\_64digits/lag\_kv\_partial\_match/qwen2.5-7b-instruct/lb\_ratio\_\_0.1667,\_lag\_size\_\_512rb.pdf}}
		\centerline{$L=512,r=6 \times$}
	\end{minipage}
	\begin{minipage}{0.24\textwidth}
		\centerline{\includegraphics[width=\textwidth]{Figures/diagrams/lagkv\_test/paulgraham\_passkey/20480words\_10x10x3\_64digits/lag\_kv\_partial\_match/qwen2.5-7b-instruct/lb\_ratio\_\_0.125,\_lag\_size\_\_512rb.pdf}}
		\centerline{$L=512,r=8 \times$}
	\end{minipage}

	\begin{minipage}{0.24\textwidth}
		\centerline{\includegraphics[width=\textwidth]{Figures/diagrams/lagkv\_test/paulgraham\_passkey/20480words\_10x10x3\_64digits/lag\_kv\_partial\_match/qwen2.5-7b-instruct/lb\_ratio\_\_0.5,\_lag\_size\_\_128rb.pdf}}
		\centerline{$L=128,r=2 \times$}
	\end{minipage}
	\begin{minipage}{0.24\textwidth}
		\centerline{\includegraphics[width=\textwidth]{Figures/diagrams/lagkv\_test/paulgraham\_passkey/20480words\_10x10x3\_64digits/lag\_kv\_partial\_match/qwen2.5-7b-instruct/lb\_ratio\_\_0.25,\_lag\_size\_\_128rb.pdf}}
		\centerline{$L=128,r=4 \times$}
	\end{minipage}
\begin{minipage}{0.24\textwidth}
		\centerline{\includegraphics[width=\textwidth]{Figures/diagrams/lagkv\_test/paulgraham\_passkey/20480words\_10x10x3\_64digits/lag\_kv\_partial\_match/qwen2.5-7b-instruct/lb\_ratio\_\_0.1667,\_lag\_size\_\_128rb.pdf}}
		\centerline{$L=128,r=6 \times$}
	\end{minipage}
	\begin{minipage}{0.24\textwidth}
		\centerline{\includegraphics[width=\textwidth]{Figures/diagrams/lagkv\_test/paulgraham\_passkey/20480words\_10x10x3\_64digits/lag\_kv\_partial\_match/qwen2.5-7b-instruct/lb\_ratio\_\_0.125,\_lag\_size\_\_128rb.pdf}}
		\centerline{$L=128,r=8 \times$}
	\end{minipage}

	\caption{The 64-digit Passkey Retrieval of Qwen-2.5-7B-Instruct for different setups with {\bf partial} matching.}
	\label{fig:qwenneedlespartial}
\end{figure*}

\begin{figure*}

	\centering
	\begin{minipage}{0.24\textwidth}
		\centerline{\includegraphics[width=\textwidth]{Figures/diagrams/lagkv\_test/paulgraham\_passkey/20480words\_10x10x3\_64digits/lag\_kv\_exact\_match/llama3.1-8b-instruct/lb\_ratio\_\_0.5rb.pdf}}
		\centerline{$L=1024,r=2 \times$}
	\end{minipage}
	\begin{minipage}{0.24\textwidth}
		\centerline{\includegraphics[width=\textwidth]{Figures/diagrams/lagkv\_test/paulgraham\_passkey/20480words\_10x10x3\_64digits/lag\_kv\_exact\_match/llama3.1-8b-instruct/lb\_ratio\_\_0.25rb.pdf}}
		\centerline{$L=1024,r=4 \times$}
	\end{minipage}
\begin{minipage}{0.24\textwidth}
		\centerline{\includegraphics[width=\textwidth]{Figures/diagrams/lagkv\_test/paulgraham\_passkey/20480words\_10x10x3\_64digits/lag\_kv\_exact\_match/llama3.1-8b-instruct/lb\_ratio\_\_0.1667rb.pdf}}
		\centerline{$L=1024,r=6 \times$}
	\end{minipage}
	\begin{minipage}{0.24\textwidth}
		\centerline{\includegraphics[width=\textwidth]{Figures/diagrams/lagkv\_test/paulgraham\_passkey/20480words\_10x10x3\_64digits/lag\_kv\_exact\_match/llama3.1-8b-instruct/lb\_ratio\_\_0.125rb.pdf}}
		\centerline{$L=1024,r=8 \times$}
	\end{minipage}

	\begin{minipage}{0.24\textwidth}
		\centerline{\includegraphics[width=\textwidth]{Figures/diagrams/lagkv\_test/paulgraham\_passkey/20480words\_10x10x3\_64digits/lag\_kv\_exact\_match/llama3.1-8b-instruct/lb\_ratio\_\_0.5,\_lag\_size\_\_512rb.pdf}}
		\centerline{$L=512,r=2 \times$}
	\end{minipage}
	\begin{minipage}{0.24\textwidth}
		\centerline{\includegraphics[width=\textwidth]{Figures/diagrams/lagkv\_test/paulgraham\_passkey/20480words\_10x10x3\_64digits/lag\_kv\_exact\_match/llama3.1-8b-instruct/lb\_ratio\_\_0.25,\_lag\_size\_\_512rb.pdf}}
		\centerline{$L=512,r=4 \times$}
	\end{minipage}
\begin{minipage}{0.24\textwidth}
		\centerline{\includegraphics[width=\textwidth]{Figures/diagrams/lagkv\_test/paulgraham\_passkey/20480words\_10x10x3\_64digits/lag\_kv\_exact\_match/llama3.1-8b-instruct/lb\_ratio\_\_0.1667,\_lag\_size\_\_512rb.pdf}}
		\centerline{$L=512,r=6 \times$}
	\end{minipage}
	\begin{minipage}{0.24\textwidth}
		\centerline{\includegraphics[width=\textwidth]{Figures/diagrams/lagkv\_test/paulgraham\_passkey/20480words\_10x10x3\_64digits/lag\_kv\_exact\_match/llama3.1-8b-instruct/lb\_ratio\_\_0.125,\_lag\_size\_\_512rb.pdf}}
		\centerline{$L=512,r=8 \times$}
	\end{minipage}

	\begin{minipage}{0.24\textwidth}
		\centerline{\includegraphics[width=\textwidth]{Figures/diagrams/lagkv\_test/paulgraham\_passkey/20480words\_10x10x3\_64digits/lag\_kv\_exact\_match/llama3.1-8b-instruct/lb\_ratio\_\_0.5,\_lag\_size\_\_128rb.pdf}}
		\centerline{$L=128,r=2 \times$}
	\end{minipage}
	\begin{minipage}{0.24\textwidth}
		\centerline{\includegraphics[width=\textwidth]{Figures/diagrams/lagkv\_test/paulgraham\_passkey/20480words\_10x10x3\_64digits/lag\_kv\_exact\_match/llama3.1-8b-instruct/lb\_ratio\_\_0.25,\_lag\_size\_\_128rb.pdf}}
		\centerline{$L=128,r=4 \times$}
	\end{minipage}
\begin{minipage}{0.24\textwidth}
		\centerline{\includegraphics[width=\textwidth]{Figures/diagrams/lagkv\_test/paulgraham\_passkey/20480words\_10x10x3\_64digits/lag\_kv\_exact\_match/llama3.1-8b-instruct/lb\_ratio\_\_0.1667,\_lag\_size\_\_128rb.pdf}}
		\centerline{$L=128,r=6 \times$}
	\end{minipage}
	\begin{minipage}{0.24\textwidth}
		\centerline{\includegraphics[width=\textwidth]{Figures/diagrams/lagkv\_test/paulgraham\_passkey/20480words\_10x10x3\_64digits/lag\_kv\_exact\_match/llama3.1-8b-instruct/lb\_ratio\_\_0.125,\_lag\_size\_\_128rb.pdf}}
		\centerline{$L=128,r=8 \times$}
	\end{minipage}

	\caption{The  64-digit Passkey Retrieval of Llama-3.1-8B-Instruct for different setups with {\bf exact} matching.}
	\label{fig:llamaneedlesexact}
\end{figure*}

\begin{figure*}

	\centering
	\begin{minipage}{0.24\textwidth}
		\centerline{\includegraphics[width=\textwidth]{Figures/diagrams/lagkv\_test/paulgraham\_passkey/20480words\_10x10x3\_64digits/lag\_kv\_exact\_match/qwen2.5-7b-instruct/lb\_ratio\_\_0.5rb.pdf}}
		\centerline{$L=1024,r=2 \times$}
	\end{minipage}
	\begin{minipage}{0.24\textwidth}
		\centerline{\includegraphics[width=\textwidth]{Figures/diagrams/lagkv\_test/paulgraham\_passkey/20480words\_10x10x3\_64digits/lag\_kv\_exact\_match/qwen2.5-7b-instruct/lb\_ratio\_\_0.25rb.pdf}}
		\centerline{$L=1024,r=4 \times$}
	\end{minipage}
\begin{minipage}{0.24\textwidth}
		\centerline{\includegraphics[width=\textwidth]{Figures/diagrams/lagkv\_test/paulgraham\_passkey/20480words\_10x10x3\_64digits/lag\_kv\_exact\_match/qwen2.5-7b-instruct/lb\_ratio\_\_0.1667rb.pdf}}
		\centerline{$L=1024,r=6 \times$}
	\end{minipage}
	\begin{minipage}{0.24\textwidth}
		\centerline{\includegraphics[width=\textwidth]{Figures/diagrams/lagkv\_test/paulgraham\_passkey/20480words\_10x10x3\_64digits/lag\_kv\_exact\_match/qwen2.5-7b-instruct/lb\_ratio\_\_0.125rb.pdf}}
		\centerline{$L=1024,r=8 \times$}
	\end{minipage}

	\begin{minipage}{0.24\textwidth}
		\centerline{\includegraphics[width=\textwidth]{Figures/diagrams/lagkv\_test/paulgraham\_passkey/20480words\_10x10x3\_64digits/lag\_kv\_exact\_match/qwen2.5-7b-instruct/lb\_ratio\_\_0.5,\_lag\_size\_\_512rb.pdf}}
		\centerline{$L=512,r=2 \times$}
	\end{minipage}
	\begin{minipage}{0.24\textwidth}
		\centerline{\includegraphics[width=\textwidth]{Figures/diagrams/lagkv\_test/paulgraham\_passkey/20480words\_10x10x3\_64digits/lag\_kv\_exact\_match/qwen2.5-7b-instruct/lb\_ratio\_\_0.25,\_lag\_size\_\_512rb.pdf}}
		\centerline{$L=512,r=4 \times$}
	\end{minipage}
\begin{minipage}{0.24\textwidth}
		\centerline{\includegraphics[width=\textwidth]{Figures/diagrams/lagkv\_test/paulgraham\_passkey/20480words\_10x10x3\_64digits/lag\_kv\_exact\_match/qwen2.5-7b-instruct/lb\_ratio\_\_0.1667,\_lag\_size\_\_512rb.pdf}}
		\centerline{$L=512,r=6 \times$}
	\end{minipage}
	\begin{minipage}{0.24\textwidth}
		\centerline{\includegraphics[width=\textwidth]{Figures/diagrams/lagkv\_test/paulgraham\_passkey/20480words\_10x10x3\_64digits/lag\_kv\_exact\_match/qwen2.5-7b-instruct/lb\_ratio\_\_0.125,\_lag\_size\_\_512rb.pdf}}
		\centerline{$L=512,r=8 \times$}
	\end{minipage}

	\begin{minipage}{0.24\textwidth}
		\centerline{\includegraphics[width=\textwidth]{Figures/diagrams/lagkv\_test/paulgraham\_passkey/20480words\_10x10x3\_64digits/lag\_kv\_exact\_match/qwen2.5-7b-instruct/lb\_ratio\_\_0.5,\_lag\_size\_\_128rb.pdf}}
		\centerline{$L=128,r=2 \times$}
	\end{minipage}
	\begin{minipage}{0.24\textwidth}
		\centerline{\includegraphics[width=\textwidth]{Figures/diagrams/lagkv\_test/paulgraham\_passkey/20480words\_10x10x3\_64digits/lag\_kv\_exact\_match/qwen2.5-7b-instruct/lb\_ratio\_\_0.25,\_lag\_size\_\_128rb.pdf}}
		\centerline{$L=128,r=4 \times$}
	\end{minipage}
\begin{minipage}{0.24\textwidth}
		\centerline{\includegraphics[width=\textwidth]{Figures/diagrams/lagkv\_test/paulgraham\_passkey/20480words\_10x10x3\_64digits/lag\_kv\_exact\_match/qwen2.5-7b-instruct/lb\_ratio\_\_0.1667,\_lag\_size\_\_128rb.pdf}}
		\centerline{$L=128,r=6 \times$}
	\end{minipage}
	\begin{minipage}{0.24\textwidth}
		\centerline{\includegraphics[width=\textwidth]{Figures/diagrams/lagkv\_test/paulgraham\_passkey/20480words\_10x10x3\_64digits/lag\_kv\_exact\_match/qwen2.5-7b-instruct/lb\_ratio\_\_0.125,\_lag\_size\_\_128rb.pdf}}
		\centerline{$L=128,r=8 \times$}
	\end{minipage}

	\caption{The 64-digit Passkey Retrieval of Qwen-2.5-7B-Instruct for different setups with {\bf exact} matching.}
	\label{fig:qwenneedlesexact}
\end{figure*}

\end{document}